%% file: main.tex
\documentclass[10pt,twocolumn,letterpaper]{article}

\usepackage[pagenumbers]{cvpr} 

\usepackage[accsupp]{axessibility} 
\usepackage{graphicx}
\usepackage{amsmath}
\usepackage{amssymb}
\usepackage{booktabs}
\usepackage{url}
\input{preamble}

\usepackage{multirow}
\usepackage{float}

\definecolor{cvprblue}{rgb}{0.21,0.49,0.74}
\usepackage[pagebackref,breaklinks,colorlinks,citecolor=cvprblue]{hyperref}

\def\paperID{5157} 
\def\confName{CVPR}
\def\confYear{2024}

\title{ReCoRe: Regularized Contrastive Representation Learning of World Model}

\author{Rudra P.K. Poudel$^1$ \qquad Harit Pandya$^1$ \qquad Stephan Liwicki$^1$\\
$^1$ Cambridge Research Laboratory\\
Toshiba Europe Ltd, UK\\
{\tt\small first-name.last-name@toshiba.eu}
\and
Roberto Cipolla$^{1,2}$\\
$^2$ Department of Engineering\\
University of Cambridge, UK\\
{\tt\small rc10001@cam.ac.uk}
}

\begin{document}
\maketitle

\begin{abstract}
While recent model-free Reinforcement Learning (RL) methods have demonstrated human-level effectiveness in gaming environments, their success in everyday tasks like visual navigation has been limited, particularly under significant appearance variations. This limitation arises from (i) poor sample efficiency and (ii) over-fitting to training scenarios. To address these challenges, we present a world model that learns invariant features using (i) contrastive unsupervised learning and (ii) an intervention-invariant regularizer. Learning an explicit representation of the world dynamics i.e. a world model, improves sample efficiency while contrastive learning implicitly enforces learning of invariant features, which improves generalization. However, the na\"ive integration of contrastive loss to world models is not good enough, as world-model-based RL methods independently optimize representation learning and agent policy. To overcome this issue, we propose an intervention-invariant regularizer in the form of an auxiliary task such as depth prediction, image denoising, image segmentation, etc.,  that explicitly enforces invariance to style interventions. Our method outperforms current state-of-the-art model-based and model-free RL methods and significantly improves on out-of-distribution point navigation tasks evaluated on the iGibson benchmark. With only visual observations, we further demonstrate that our approach outperforms recent language-guided foundation models for point navigation, which is essential for deployment on robots with limited computation capabilities. Finally, we demonstrate that our proposed model excels at the sim-to-real transfer of its perception module on the Gibson benchmark. 
\end{abstract}



\section{Introduction}
In recent years, deep RL algorithms have been successfully employed for designing optimal strategies for games \cite{rl-atari-mnih13,alpha-zero-silver18} and shown a promise for controlling robots \cite{rl-control-levine16,ddppo-wijmans19}. Model-free RL approaches learn the policy along with the visual encoder in an end-to-end fashion from raw observations. As a result, they require a large number of training samples, which makes it difficult to deploy them on real robots where obtaining a large amount of training data is resource intensive, especially for safety-critical tasks such as autonomous navigation \cite{d4rl-gurtler23}. On the contrary, in model-based RL, an explicit predictive model of the world is learned called \emph{world model}, enabling the agent to plan by thinking ahead \cite{pilco-deisenroth2011,alpha-zero-silver18,wm-ha18,dreamerv2-hafner21}. The world model is learned separately from the policy, therefore, the policy can use the world model as a surrogate for the real world.  Consequently, model-based methods have higher sample efficiency \cite{wm-ha18,dreamerv2-hafner21} making them more suitable in real environments since they can be trained with a small amount of data.
\begin{figure}[t]
\begin{center}
\includegraphics[width=0.7\linewidth]{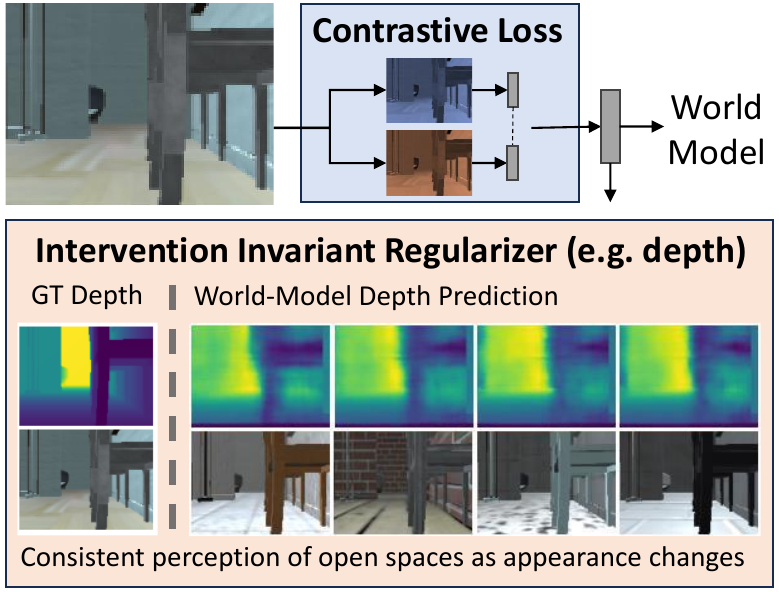}
\end{center}
\caption{Intervention invariant regularizer is applied in addition to the contrastive loss in world-model-based RL. ReCoRe learns robust representations that are invariant to out-of-distribution appearance variations which help in the generalization of downstream tasks (such as navigation). Notice the consistent depth predictions despite texture variations of an iGibson evaluation scene.}
\label{fig:teaser}
\end{figure}

Nevertheless, even current model-based RL struggles with generalization, as model-based approaches have to learn the world model purely from experience, which poses several challenges: The central issue is the training bias, which can be exploited by an agent, and leads to poor performance when deployed \cite{wm-ha18}. Another issue is that the latent representation is learned from a reconstruction loss, such as the state abstraction of variational autoencoders (VAE) \cite{vae-kingma2014}, which is not sufficient to separate the task-relevant states from irrelevant ones. Hence, the RL policy may still overfit to environment-specific characteristics \cite{block-mdp-zhang20}. Thus, the aim of this paper is invariant feature abstraction, which is essential for learning a robust RL policy.

\begin{figure*}[t]
\begin{center}
\includegraphics[width=1.0\linewidth]{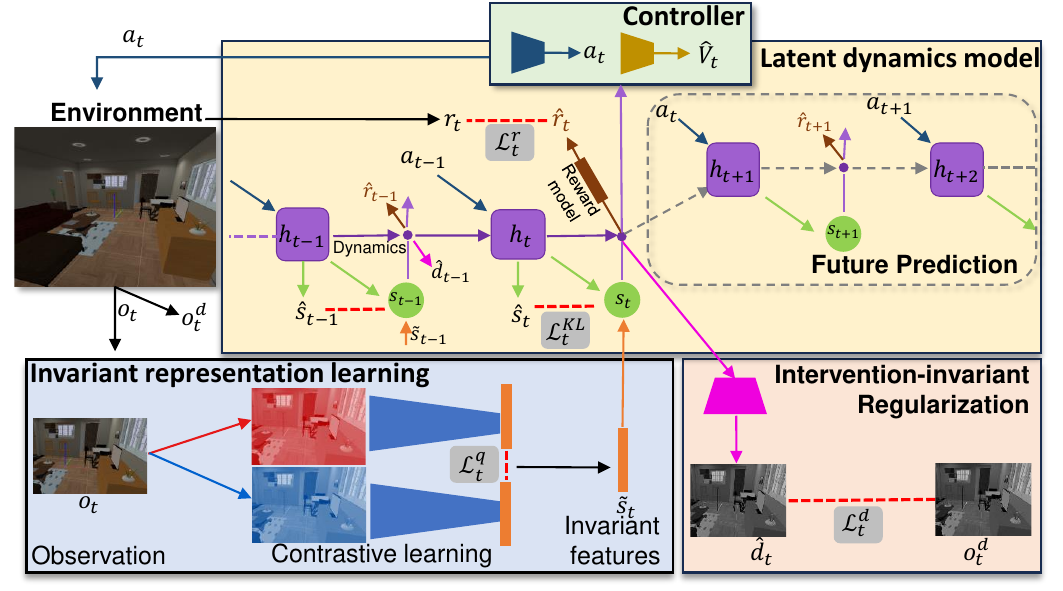}
\end{center}
\caption{Flow diagram of proposed \textit{Regularized Contrastive Representation learning} (ReCoRe) of World Model. It consists of four components: (i) invariant representation learning module, (ii) intervention-invariant regularizer, (iii) latent dynamics model, and (iv) actor-critic controller. The invariant representation learning module utilizes data augmentation and contrastive learning for invariant features abstraction ($\tilde{s}_t$)  from image observations ($o_t$). The latent dynamics model employs a recurrent neural network with deterministic hidden states ($h_t$) to predict the stochastic latent prior states ($\widehat{s}_t$), and corresponding rewards ($\widehat{r}_t$) from the posterior ($s_t$). Intervention invariant regularizer considers an auxiliary task (here depth prediction i.e. $p_{\theta}(\widehat{d}_t \vert s_t,h_t)$) invariant to data augmentation (here texture variations) which prevents feature collapse in training the world-model with contrastive learning. The controller maximizes the expected rewards of the action using an actor critic approach. In addition to being sample efficient, the proposed approach is more robust to out-of-distribution and sim-to-real generalization, since the controller is learned separately using invariant states of the environment.
}
\label{fig:ReCoRe}
\end{figure*}

Specifically, we propose to use contrastive learning for invariant state abstraction since the contrastive learning objective implicitly ensures that the feature embeddings are invariant to the intervention, i.e. data augmentation. However, feature collapse is possible if a na\"ive implementation is used as can be seen in our evaluation (Table \ref{tbl:igibson1-results}), where the na\"ive implementation of contrastive loss (ReCoRe-D) completely fails ($<$1\% success rate) for model-based RL. Mitrovic \etal \cite{invariant-causal-mechanisms-mitrovic21} proposed a regularizer based on KL-divergence that matches the distributions among the augmentations, which stabilizes contrastive learning under model-free RL settings and slightly improved the performance over CURL~\cite{curl-laskin20}.  On the contrary, we propose a regularizer in the form of an auxiliary task to explicitly enforce the invariant feature learning. For example in the navigation task, we utilize depth predictions to extract the geometric features needed for navigation as they do not depend on textures as shown in Figure \ref{fig:teaser}. We emphasize that depth is only required for training but not for deployment since it works as a regulariser. Furthermore, in cases where depth is not available other auxiliary tasks such as image denoising, segmentation or optical flow prediction can be utilized for regularization, enabling a wider applicability of the proposed model. Importantly, our setup allows us to employ contrastive learning in model-based RL settings, which improves the sample efficiency and helps with Out-of-Distribution (OoD) generalization. 

In summary, we propose a \textit{Regularized Contrastive Representation learning} (ReCoRe) approach to the world model. The proposed variant of the world model can extract and predict robust invariant features (Figure \ref{fig:ReCoRe}). ReCoRe is verified on the \textit{point goal} navigation task from Gibson \cite{gibson-xia} and iGibson 1.0 \cite{igibson1-shen21} as well as on the DeepMind Control suite (DMControl) \cite{dmc-tunyasuvunakool2020}.  
Thus, our main contributions are:
\begin{enumerate}
    \item We show that contrastive unsupervised representation learning can significantly improve OoD generalization of world model based reinforcement learning (Table  \ref{tbl:igibson1-results}, ReCoRe vs. DreamerV2).
    \item We propose an intervention-invariant regularizer that learns the invariant features (Section \ref{sec:invariant_encoder}) and is shown to be crucial in preventing feature collapse of contrastive learning (Table \ref{tbl:igibson1-results}, ReCoRe vs. ReCoRe-D).
    \item Through extensive experiments, we showcase that our approach outperforms state-of-the-art RL models (including language guided foundation model Grounding DINO \cite{gdino-liu23}) on out-of-distribution generalization (Table \ref{tbl:igibson1-results}) and sim-to-real transfer of learned features (Table \ref{tbl:sim2real-gibson-results}). We further show that even for in-distribution evaluation our approach outperforms model-free reinforcement learning approaches (Table \ref{tbl:dmc-results}) which is difficult for other model-based learning approaches. 
    
\end{enumerate}

\section{Related Work}

\textbf{Unsupervised Representation Learning.}
Learning reusable feature representations from large unlabeled data is a fundamental challenge in machine learning. In the context of computer vision, one can leverage unlabeled images and videos to learn good intermediate representations, which can be useful for a wide variety of downstream tasks. 
Recently, VAE \cite{vae-kingma2014} has been a preferred approach for representation learning in model-based RL \cite{wm-ha18}. Since VAE does not make any additional consideration of downstream tasks, invariant representation learning with contrastive loss has shown more promising results \cite{anand19,curl-laskin20}.  
Self-supervised learning formulates representation learning as a supervised loss function between different transformations of data. In image-based learning self-supervision can be formulated using different image augmentations, for example, image distortion and rotation \cite{ssl-dosovitskiy14,simclr-chen20}. We also use different data augmentation techniques to learn the invariant features using contrastive loss. Recently, transformer-based visual models such as DINO \cite{dino-caron23} have been shown to intrinsically capture robust object representations through self-supervision. Combining such models with language models pretrained on a large corpus of data \cite{lseg-li22,gdino-liu23} have resulted in powerful representations that generalize well to diverse environments. However, inference on such large models is computationally expensive which makes it difficult to deploy on low-budget robots. We also compare our proposed approach to Grounding DINO \cite{gdino-liu23} features, which we believe is the strongest baseline, and showcase superior results.    

\textbf{Contrastive Learning.}
Representation learning methods based on contrastive loss  \cite{closs-chopra05} have achieved state-of-the-art performance on face verification tasks. These methods use a contrastive loss to learn representations invariant to data augmentation \cite{simclr-chen20,moco_he20}. Given a list of input samples, contrastive loss forces samples from the same class to have similar embeddings and different ones for different classes. Since class labels are not available in the unsupervised setting, contrastive loss forces similar embedding for the augmented version of the same sample and different ones for different samples.  
There are several ways of formulating the contrastive loss such as Siamese \cite{closs-chopra05}, InfoNCE \cite{info-nce-oord18}, and SimCLR \cite{simclr-chen20}. In this work, we chose InfoNCE \cite{info-nce-oord18} for our contrastive loss

\textbf{Learning Invariant Features.}
Learning structured representations that capture the underlying causal mechanisms generating the data is a central problem for robust machine learning systems \cite{crl_scholkopf21}. However, recovering the underlying causal structure of the environment from observational data without additional assumptions is a complex problem. A recent successful approach for causal discovery, in the context of unknown causal structure, is causal inference using invariant prediction \cite{causalilty-invariance-peters16}. 
Mitrovic \etal \cite{invariant-causal-mechanisms-mitrovic21} recently formalized self-supervised representation learning using invariant causal mechanisms. Our proposed world model also exploits the invariance principle, which is formalized using contrastive loss to learn invariant features. In section \ref{sec:proposed-model}, we explain how we utilized the data augmentation technique to learn the invariant state of the environment.

\textbf{Model-based RL.}
The human brain discovers the underlying hidden causes of an observation. Those internal representations of the world influence how agents infer which actions will lead to a higher reward \cite{survey-mental-simulation-hamrick2019}. An early example of this idea was put forward by Sutton \cite{dyna-sutton90}, where future hallucination samples rolled out from the learned world model are used in addition to the agent's interactions for sample efficient learning. Further, planning through the world model has been successfully demonstrated in the \textit{world model} by Ha \etal \cite{wm-ha18} and DreamerV2 by Hafner \etal \cite{dreamerv2-hafner21}. Recently, replacing state extraction \cite{mwm-seo23} and dynamic prediction \cite{twm-robine2023} using transformer architecture is a popular direction, which further improves the results \cite{mwm-seo23}. Masked World Models (MWM) combines the transformer-based masked autoencoder with DreamerV2. Other Dreamer variants DayDreamer \cite{daydreamer-wu23} and DreamerV3 \cite{dreamerv3-hafner23} aim to scale up DreamerV2 architecture for physical robots and Atari games. Specifically, DayDreamer \cite{daydreamer-wu23} trains DreamerV2 on physical robots, while DreamerV3 \cite{dreamerv3-hafner23} proposes design choices (e.g. symlog scaling and EMA regularization) that help in scaling up DreamerV2 to several domains. In our work, we propose to learn invariant features to improve OoD generalization and sample efficiency further. While we utilize a similar architecture as DreamerV2 for policy and world model, our method is complementary to DreamerV3 and can take advantage of their design choices.   

\textbf{Sample Efficiency.}
Joint learning of auxiliary tasks with model-free RL makes them competitive with model-based RL in terms of sample efficiency. For example, the recently proposed model-free RL method called CURL \cite{curl-laskin20} added contrastive loss as an auxiliary task and outperformed the state-of-the-art model-based RL method called Dreamer \cite{dreamer_hafner20}. Also, two recent works using data augmentation for RL called RAD \cite{data-aug-laskin20} and DrQ \cite{data-aug-yarats21} outperform CURL without using an auxiliary contrastive loss. These results warrant that if an agent has access to a rich stream of data from the environments, an additional regularizer is unnecessary since directly optimizing the policy objective is better than optimizing multiple objectives. However, we do not have access to a rich stream of data for many complex problems, hence sample efficiency still matters. Further, these papers do not consider the effect of regularizers in the form of auxiliary tasks and unsupervised representation learning for model-based RL, which is the main focus of our work.

\section{ReCoRe-based World Model}
\label{sec:proposed-model}

We consider the visual control task as a finite-horizon partially observable Markov decision process (POMDP). We denote observation space, action space and time horizon as $\mathcal{O}$, $\mathcal{A}$ and $\mathcal{T}$ respectively. An agent performs continuous actions $a_t \sim p(a_t|o_{\leq t},a_{<t})$, and receives observations and scalar rewards $o_t, r_t \sim p(o_t, r_t | o_{<t}, a_{<t})$ from the unknown environment. 
The goal of an agent is to maximize the expected total rewards $E_p(\sum_{t=1}^{T} r_t)$. In the following sections, we detail our proposed model. 

\subsection{World Model Design}
We propose our \textit{Regularized Contrastive Representation learning} (ReCoRe) technique to learn the world model. The data flow diagram of our proposed world model with explicitly regularized invariant feature learning is shown in Figure \ref{fig:ReCoRe}. Our method consists of four main components: (i) invariant representation learning module,  (ii) intervention-invariant regularizer, (iii) latent dynamics model, and (iv) the controller. 
Next, we describe the components in detail. 

\subsubsection{Invariant Representations Learning Module}
\label{sec:invariant_encoder}
Extracting representations that are invariant to appearance variations from image observations is a key component of our model. This is crucial for improving robustness and out-of-distribution generalization of the model in the real world.  We learn these invariant representations by maximizing agreement between different style interventions of the same observation via a contrastive loss in the latent feature space. The world model optimizes feature learning and controller separately to improve the sample efficiency and simplify controller learning. 
Motivated by the fact that most of the complexity of model-based RL approaches resides in the world model (i.e. the feature extraction and the dynamics model), we hypothesize that an additional supervisory signal from an auxiliary task helps to learn a better state representation. 
In this work, we used InfoNCE \cite{info-nce-oord18} style loss to learn invariant features. Hence, our encoder takes RGB observations ($o_t$) as inputs and decodes rewards ($r_t$) as well as depth ($o_t^d$) as auxiliary outputs during the training phase. 
The invariant feature learning is enforced by contrastive loss ($\mathcal{L}^q_t$). 
The proposed invariant features learning technique has the following three sub-modules:
\begin{itemize}
    \item A \textit{style intervention} module that utilizes data augmentation. We use spatial jitter, Gaussian blur, color jitter, grayscale and cutout data augmentation techniques for style intervention. Spatial jitter is implemented by first padding and then performing random crop. 
    Given any observation $o_t$, our style intervention module randomly transforms it into two correlated views of the same observations, used for contrastive learning. All the hyperparameters are provided in the appendix.
    
    \item We use an \textit{encoder} network that extracts representations from augmented observations, which is $\Tilde{s_t} = encoder(o_t)$. The encoder is optimized using contrastive loss (Equation \ref{eq:info-nce-loss}), which by construction makes the encoder invariant to the augmentations. We additionally employ an EMA regularization \cite{moco_he20} on the encoder for stabilizing training. The hyperparameters of the encoder are provided in the appendix.  
    
    \item \textit{Contrastive loss} is defined for a contrastive prediction task, which can be explained as a differentiable dictionary lookup task. Given a query observation $q$ and a set of keys $K = \{k_0,k_1,...k_{2B}\}$ of length $2B$, with known positive $\{k_+\}$ and negative $\{k_-\}$ keys, the aim of contrastive loss is to learn a representation in which positive sample pairs stay close to each other while negative ones are far apart. In contrastive learning $q$, $K$, $k_+$ and ${k_{-}}$ are also known as \textit{anchors}, \textit{targets}, \textit{positive} and \textit{negative} samples. We use bilinear products for \textit{projection head} $W$ and InfoNCE loss for contrastive learning \cite{info-nce-oord18}, which enforces the desired similarity in the embedding space:
    \end{itemize}
    \begin{equation} \label{eq:info-nce-loss}
        \mathcal{L}^q_t = -\log\frac{\exp(q^TWk_+)}{\exp(q^TWk_+) + \sum_{i=0}^{2(B-1)}\exp(q^TWk_i)}
    \end{equation}

\subsubsection{Intervention-invariant Regularizer}
Model-based RL approaches such as DreamerV2, separate the training of the world model and controller. This makes model-based RL more sample efficient as compared to model-free RL, since future predictions from latent dynamics (rollouts) can be used to learn RL policy. However, the policy learning is disconnected from world-model learning. Therefore, the supervisory signal from the policy does not optimize the encoder. While the image reconstruction loss in DreamerV2 caters to this issue for in-distribution learning, it is not enough to overcome the out-of-distribution generalization. Contrastive learning further elevates the issue, since it trains the encoder under different data augmentations while the world model still aims to predict the unaugmented image using latent dynamics. This leads to the collapse of encoder features due to a lack of correct supervisory signals. Therefore the brute-force combination of world-model and contrastive loss fails (as can be seen in Table \ref{tbl:igibson1-results}, ReCoRe vs. ReCoRe-D).  
To this end, we propose regularization in the form of an intervention invariant auxiliary task to explicitly enforce invariance, which is robust against changes to the nuisance variables. For navigation tasks, we choose depth reconstruction $p_{\theta}(\widehat{d}_t \vert s_t,h_t)$ to verify our proposal since geometrical information remains invariant to appearance variations. The regularization task enforces the latent state to predict consistent pixel-wise depth under image augmentations by minimizing the negative of log-likelihood loss ($\mathcal{L}_t^d=-\ln q(o^d_t|s_t,h_t)$). We also experiment with image denoising and semantic segmentation on DeepMind control suite \cite{dmc-tunyasuvunakool2020}, where depth is not available. However, denoising is not truly invariant to RGB observation hence the performance gain on DeepMind control suite is limited but dense semantic segmentation improves the results.  
Other examples of intervention invariant auxiliary tasks for our augmentations are dense scene flow or sparse landmarks detection. Furthermore, our idea paves the foundation and generalizes easily to other interventions for future tasks. The key challenge is to design different interventions and intervention-invariant auxiliary tasks.

\subsubsection{Learning Latent Dynamics}
We leverage the DreamerV2 approach to train our world model. Similar to \cite{dreamerv2-hafner21}, the latent dynamics are modeled as a recurrent state space model which relies on recurrent neural network $f_{\theta}$ that utilizes its deterministic hidden state $h_t$ to predict the prior stochastic state $\widehat{s}_{t}$. This enables efficient latent imagination for planning \cite{wm-ha18,dreamer_hafner20}. Thus, dynamics models and representation learning modules are tightly integrated as world model and have the following components:

\begin{gather}
\begin{aligned}
 & \text{Recurrent model:} && h_t = f_{\theta}(h_{t-1}, s_{t-1}, a_{t-1}) \\
 & \text{Representation model:} && p_\theta(s_t|h_t,\Tilde{s_t}) \\
 & \text{Reward prediction model:} && q_\theta(\widehat{r}_t|s_t,h_t) \\
 & \text{Latent dynamics model:} && q_\theta(\widehat{s}_t|h_t).
\end{aligned}
\end{gather}

The world model is represented by neural networks and $\theta$ represents their combined parameters. It is  optimized jointly with the invariant representation learning module and intervention-invariant regularizer, by minimizing,
 \begin{equation}
     \mathcal{L}_{WM} =  E_p \left ( \sum_{t} \left ( \mathcal{L}^q_t + \mathcal{L}_t^d + \mathcal{L}_t^r  + \beta \mathcal{L}^{KL}_t \right ) \right )
 \end{equation}
where, $\mathcal{L}_t^d=-\ln q(o^d_t|s_t)$, $\mathcal{L}_t^r =-\ln q(r_t|s_t)$ and \mbox{$\mathcal{L}^{KL}_t = KL(p(s_t|s_{t-1},a_{t-1},\Tilde{s_t}) || q(s_t|s_{t-1},a_{t-1}))$}.

\subsubsection{Learning Controller}
The objective of the controller is to optimize the expected rewards of the action, which is optimized using an actor critic approach. The actor critic approach considers the rewards beyond the horizon. Inspired by world model \cite{wm-ha18} and Dreamer \cite{dreamer_hafner20}, we learn an action model and a value model in the imagined latent space of the world model. The action model implements a policy that aims to predict future actions that maximizes the total expected rewards in the imagined environment. Given $H$ as the imagination horizon length and $\gamma$ as the discount factor for the future rewards, the action and value model are defined as follows:
\begin{gather}
\begin{aligned}
 & \text{Action model:} && q_\phi(\widehat{a}_t|\widehat{s}_{t}) \\
 & \text{Value model:} && E_{q(\cdot|\widehat{s}_\tau)}{\textstyle \sum_{\tau=t}^{t+H}\gamma^{\tau-t}\widehat{r}_\tau}.
\end{aligned}
\end{gather}


\subsection{Implementation Details}
The proposed ReCoRe expands on the publicly available code base of DreamerV2 \cite{dreamerv2-hafner21}. Following MoCo \cite{moco_he20} and BYOL \cite{byol-grill20} we have used the moving average version of the query encoder to encode the keys $K$ with a momentum value of 0.999. The contrastive loss is jointly optimized with the world model using Adam \cite{adam-kingma2014}. We have used five layers encoder with a starting number of feature maps equal to 32, then doubled in every consecutive layer. 
To encode the task observations we used two dense layers of size 32 with ELU activations \cite{elu-clevert2015}. The features from RGB image observation and task observation are concatenated before sending to the representation model. Replay buffer capacity is $3e^5$ for both 100k and 500k steps experiments. We update the model parameters on every fifth interactive step. Further, all architectural details and hyperparameters are provided in the appendix. The training time of ReCoRe is around 3 days on a workstation with two Nvidia GeForce RTX 3090 for 500k steps, which is twice higher than the closest state-of-the-art model-based RL model DreamerV2 \cite{dreamerv2-hafner21}.

\begin{table*}[t]
\begin{center}
\begin{tabular}{lcrrrrrr|rr}
\hline
\noalign{\smallskip}
\multicolumn{2}{c}{} & \multicolumn{2}{c}{Ihlen\_0\_int} & \multicolumn{2}{c}{Ihlen\_1\_int} & \multicolumn{2}{c}{Rs\_int} & \multicolumn{2}{c}{\textbf{Env Avg}}\\
\noalign{\smallskip}
Models & Steps & \multicolumn{1}{c}{SR} &  \multicolumn{1}{c}{SPL} & \multicolumn{1}{c}{SR} &  \multicolumn{1}{c}{SPL} & \multicolumn{1}{c}{SR} &  \multicolumn{1}{c}{SPL} & \multicolumn{1}{c}{SR} &  \multicolumn{1}{c}{SPL}\\
\noalign{\smallskip}
\hline
\noalign{\smallskip}
RAD & 100k & 0.6 & 0.01 & 0.1 & 0.00 & 0.8 & 0.01 & 0.5 & 0.01\\
CURL & 100k & 8.0 & 0.07 & 0.6 & 0.01 & 5.4 & 0.05 & 4.7 & 0.04\\
MWM & 100k & 1.6 & 0.01 & 0.5 & 0.00 & 2.9 & 0.02 & 1.7 & 0.01\\
DreamerV2 & 100k & 1.8 & 0.01 & 0.6 & 0.00 & 1.7 & 0.01 & 1.3 & 0.01\\
DreamerV2 + DA & 100k & 7.3 & 0.05 & 1.6 & 0.01 & 7.7 & 0.05 & 5.5 & 0.04\\
DV2+G DINO & 100k & \textbf{48.9} & \textbf{0.45} & \textbf{17.0} & \textbf{0.14} & 45.4 & 0.38 & \textbf{37.1} & \textbf{0.33}\\
ReCoRe & 100k & 44.8 & 0.38 & 12.2 & 0.09 & \textbf{50.9} & \textbf{0.41} & 36.0 & 0.29\\
\hline
ReCoRe - CL & 100k & 1.0 & 0.01 & 0.5 & 0.00 & 2.3 & 0.01 & 1.3 & 0.01\\
ReCoRe - CL + DA & 100k & 5.2 & 0.03 & 1.3 & 0.01 & 8.3 & 0.05 & 4.9 & 0.03\\
ReCoRe - D & 100k & 0.1 & 0.00 & 0.0 & 0.00 & 0.0 & 0.00 & 0.0 & 0.00\\
ReCoRe - D + I & 100k & 15.4 & 0.12 & 4.6 & 0.04 & 17.1 & 0.12 & 12.4 & 0.09\\
\noalign{\smallskip}
\hline
\hline
\noalign{\smallskip}
RAD & 500k & 48.8 & 0.44 & 11.6 & 0.11 & 48.5 & 0.44 & 36.3 & 0.33\\
CURL & 500k & 40.8 & 0.37 & 11.4 & 0.10 & 41.9 & 0.36 & 31.4 & 0.28\\
MWM & 500k & 2.6 & 0.02 & 0.7 & 0.00 & 4.2 & 0.02 & 2.5 & 0.01\\
DreamerV2 & 500k & 1.3 & 0.01 & 0.8 & 0.01 & 2.3 & 0.02 & 1.5 & 0.01\\
DreamerV2 + DA & 500k & 14.7 & 0.10 & 3.9 & 0.03 & 20.5 & 0.13 & 13.0 & 0.08\\
DV2+G DINO & 500k & 60.9 & 0.58 & 23.2 & 0.19 & 65.8 & 0.58 & 50.0 & 0.45\\
ReCoRe & 500k & \textbf{75.3} & \textbf{0.65} & \textbf{26.5} & \textbf{0.20} & \textbf{77.3} & \textbf{0.65} & \textbf{59.7} & \textbf{0.50}\\
\hline
ReCoRe - CL & 500k & 5.5 & 0.04 & 1.4 & 0.01 & 8.0 & 0.05 & 5.0 & 0.03\\
ReCoRe - CL + DA & 500k & 27.1 & 0.19 & 7.8 & 0.05 & 31.5 & 0.22 & 22.1 & 0.16\\
ReCoRe - D & 500k & 0.9 & 0.01 & 0.2 & 0.00 & 1.2 & 0.01 & 0.8 & 0.01\\
ReCoRe - D + I & 500k & 28.6 & 0.21 & 6.6 & 0.05 & 22.3 & 0.15 & 19.2 & 0.13\\
\hline
\end{tabular}
\end{center}
\caption{Out-of-distribution generalization results on PointGoal navigation task from iGibson 1.0 dataset. \textit{Success rate} (SR) and \textit{Success weighted by (normalized inverse) Path Length} (SPL) are shown. We have trained on five scenes, and tested on held-out three scenes and visual textures. Our proposed world model with ReCoRe outperforms state-of-the-art RL models RAD, CURL, DreamerV2 and MWM and DV2+G DINO on 500K interactive steps, while it is on par with DV2+G DINO on 100K steps. Even though \textit{data augmentation} (DA) improves the DreamerV2, the proposed invariant features learning technique with \textit{contrastive loss} (CL) and \textit{intervention invariant auxiliary task} (D) is significantly better. ReCoRe collapses when we remove the auxiliary D, however replacing with common \textit{RGB image} reconstruction (I) recovers the performance slightly. Where, $-$ denotes `without' and $+$ denotes `with'.
}
\label{tbl:igibson1-results}
\end{table*}

\section{Experiments}
In our evaluation, we test ReCoRe on 3 datasets and compare it to the state of the art in model-based and model-free RL. Specifically, we use the \textit{PointGoal} navigation task from the iGibson 1.0 environment \cite{igibson1-shen21} to evaluate out-of-distribution (OoD) generalization, and include the Gibson dataset \cite{gibson-xia} to test sim-to-real performance. Here we follow common practice as we compare performance based on \textit{Success Rate} (SR) and \textit{Success weighted by (normalized inverse) Path Length} (SPL) at 100k and 500k environment steps, which tests sample efficiency \cite{curl-laskin20,data-aug-laskin20,data-aug-yarats21,dreamerv2-hafner21}. We further report on results for the DMControl suite \cite{dmc-tunyasuvunakool2020}, and present an ablation study.

\subsection{Baselines}
We compare our approach against state-of-the-art model-free RL methods RAD \cite{data-aug-laskin20} and CURL \cite{curl-laskin20}, model-based RL method DreamerV2 \cite{dreamerv2-hafner21}, recent transformer based Masked World Model (MWM) \cite{mwm-seo23} and language guided foundation model Grounding DINO \cite{gdino-liu23}. We train RAD, CURL, DreamerV2 and MWM from scratch with the respective hyperparameters proposed by the authors using the official source code provided. For Grounding DINO we freeze the visual encoder and train the world model and policy (DreamerV2) from scratch using the features from the visual encoder, which we refer to as DV2+G DINO. We believe that DV2+G DINO is the strongest baseline since it has been pretrained on a large amount of data, furthermore language guidance makes the representation highly robust.  Through our experiments in Table \ref{tbl:igibson1-results},\ref{tbl:sim2real-gibson-results} and \ref{tbl:dmc-results}, it can be seen that (i) RAD simply overfits to the in-distribution data, though after observing large amounts of data (500K environment steps) its performance improves over CURL.  (ii) DreamerV2 lacks any data augmentation in the encoder like CURL/RAD, therefore it performs significantly worse for OoD generalization.  (iii) MWM relies on masking that forces the encoder to learn missing data, therefore it performs slightly better than DreamerV2. However, masking only helps in interpolation and not in extrapolation, consequently the OoD generalization is poor. (iv) Being a pretrained model DV2+G DINO shows high sample efficiency for a smaller amount of data and is on par with our approach. However, after observing a sufficient amount of data our approach shows significantly better performance.   

\begin{table*}[t]
\begin{center}
\begin{tabular}{lcrrrrrrrrrrrr}
\hline
\noalign{\smallskip}
\multicolumn{2}{c}{} & \multicolumn{2}{c}{Ihlen} & \multicolumn{2}{c}{Muleshoe} & \multicolumn{2}{c}{Uvalda} & \multicolumn{2}{c}{Noxapater} & \multicolumn{2}{c}{McDade} & \multicolumn{2}{c}{\textbf{Env Avg}}\\
\noalign{\smallskip}
Models & Steps & \multicolumn{1}{c}{SR} &  \multicolumn{1}{c}{SPL} &  \multicolumn{1}{c}{SR} &  \multicolumn{1}{c}{SPL} &  \multicolumn{1}{c}{SR} &  \multicolumn{1}{c}{SPL} &  \multicolumn{1}{c}{SR} &  \multicolumn{1}{c}{SPL} &  \multicolumn{1}{c}{SR} &  \multicolumn{1}{c}{SPL} &  \multicolumn{1}{c}{SR} &  \multicolumn{1}{c}{SPL}\\
\noalign{\smallskip}
\hline
\noalign{\smallskip}
RAD & 100k & 0.0 & 0.00 & 0.0 & 0.00 & 0.0 & 0.00 & 0.0 & 0.00 & 0.0 & 0.00 & 0.0 & 0.00\\
CURL & 100k & 5.9 & 0.05 & 3.8 & 0.03 & 5.1 & 0.04 & 5.9 & 0.05 & 12.8 & 0.11 & 6.7 & 0.06\\
DV2 + G DINO & 100k & 38.0 & \textbf{0.34} & 33.5 & 0.29 & 39.7 & \textbf{0.35} & 37.7 & \textbf{0.33} & \textbf{55.1} & \textbf{0.51} & 40.8 & \textbf{0.36}\\
ReCoRe & 100k & \textbf{39.1} & 0.32 & \textbf{38.6} & \textbf{0.31} & \textbf{40.9} & 0.32 & \textbf{41.1} & \textbf{0.33} & 48.2 & 0.39 & \textbf{41.6} & 0.33\\
\noalign{\smallskip}
\hline
\noalign{\smallskip}
RAD & 500k & 26.4 & 0.23 & 27.5 & 0.24 & 28.5 & 0.25 & 28.6 & 0.25 & 40.0 & 0.34 & 30.2 & 0.26\\
CURL & 500k & 36.8 & 0.33 & 29.3 & 0.27 & 33.7 & 0.30 & 35.2 & 0.32 & 53.8 & 0.50 & 36.7 & 0.33\\
DV2 + G DINO & 500k & 60.2 & 0.56 & 58.0 & 0.53 & 58.9 & 0.54 & 58.5 & 0.54 & 66.2 & \textbf{0.64} & 60.3 & 0.56\\
ReCoRe & 500k & \textbf{74.3} & \textbf{0.64} & \textbf{72.9} & \textbf{0.62} & \textbf{73.5} & \textbf{0.61} & \textbf{72.0} & \textbf{0.61} & \textbf{67.0} & 0.57 & \textbf{71.9} & \textbf{0.61}\\
\hline
\end{tabular}
\end{center}
\caption{iGibson-to-Gibson dataset: sim-to-real perception transfer results on navigation task. We choose \textit{success rate} (SR) and \textit{Success weighted by (normalized inverse) Path Length} (SPL) for evaluation of the models. We have trained the models on the artist created textures of iGibson, and tested on five held-out scenes from the Gibson Dataset, which are 3D scan of the real scenes. Our proposed ReCoRe outperforms state-of-the-art RL models RAD, CURL and even model-based RL combined with language guided foundation models (DV2+G DINO) on 100k and 500k interactive steps. 
}
\label{tbl:sim2real-gibson-results}
\end{table*}

\subsection{Out-of-Distribution Generalization}
We have tested our proposed ReCoRe on random \textit{PointGoal} navigation tasks of the iGibson 1.0 environment \cite{igibson1-shen21} for OoD generalization. The iGibson dataset contains 15 floor scenes with 108 rooms. The scenes are replicas of real-world homes with artist designed textures and materials. RGB, depth and task related observation (\ie goal location, current location, and linear and angular velocities of the robot) are used. We emphasize, depth is only used during training. Actions include rotation in radians and forward distance in meters for the TurtleBot. We split iGibson for OoD generalization, as we chose five scenes for training and tested on the held-out three scenes. We also held out visual textures for all object classes. The details are provided in the appendix.

\begin{table*}[t]
\begin{center}
\begin{tabular}{lrrrr|rrr}
\hline
\noalign{\smallskip}
100k Steps Total Rewards & ReCoRe+I & ReCoRe+S & CURL & Dreamer & RAD & SAC+AE & Pixel SAC\\
\noalign{\smallskip}
\hline
\noalign{\smallskip}
Finger, spin    & 486$\pm$191 & 474$\pm$53 &  \textbf{767$\pm$56 } & 341$\pm$70 & 856$\pm$73 &740$\pm$64 &  179$\pm$66 \\
Cartpole, swingup & 472$\pm$67 & 449$\pm$121 & \textbf{582$\pm$146} & 326$\pm$27 & 828$\pm$27 & 311$\pm$11 & 419$\pm$40 \\
Reacher, easy    & 327$\pm$98 & \textbf{982$\pm$9} & 538$\pm$233 & 314$\pm$155 & 826$\pm$219 & 274$\pm$14 & 145$\pm$30 \\
Cheetah, run   & 321$\pm$78 & \textbf{400$\pm$56} & 299 $\pm$48 & 235$\pm$ 137 & 447$\pm$88 & 267$\pm$24 &  197$\pm$15 \\
Walker, walk      & 654$\pm$100 & \textbf{739$\pm$133} & 403$\pm$24 &  277$\pm$12 & 504$\pm$191 & 394$\pm$22 & 42$\pm$12 \\
Ball in cup, catch  & 830$\pm$118 & \textbf{859$\pm$287} & 769$\pm$43 & 246$\pm$174 & 840$\pm$179 & 391$\pm$82 & 312$\pm$63 \\ 
\noalign{\smallskip}
\hline
\noalign{\smallskip}
500K Steps Total Rewards &  &  &  &  &  &  \\
\noalign{\smallskip}
\hline
\noalign{\smallskip}
Finger, spin    & 471$\pm$173 & 591$\pm$181 & \textbf{ 926$\pm$45} & 796$\pm$183  & 947$\pm$101 & 884$\pm$128 &   179$\pm$166 \\
Cartpole, swingup & 675$\pm$64 & 777$\pm$64 & \textbf{841$\pm$45}& 762$\pm$27 & 863$\pm$9 & 735$\pm$63 & 419$\pm$40 \\
Reacher, easy    & 891$\pm$72 & \textbf{955$\pm$38} & 929$\pm$44 & 793$\pm$164 & 955$\pm$71 & 627$\pm$58 & 145$\pm$30 \\
Cheetah, run   & 633$\pm$70 & \textbf{731$\pm$51} & 518$\pm$28 & 570$\pm$253 & 728$\pm$71 & 550$\pm$34 &  197$\pm$15 \\
Walker, walk      & \textbf{965$\pm$4} & 960$\pm$2 & 902$\pm$43 & 897$\pm$49 & 918$\pm$16 & 847$\pm$48  & 42$\pm$12 \\
Ball in cup, catch  & 950$\pm$20 & \textbf{984$\pm$5} & 959$\pm$27 & 879$\pm$ 87 & 974$\pm$12 & 794$\pm$  58  &  312$ \pm$  63 \\
\hline
\end{tabular}
\end{center}
\caption{Experiment results on DMControl. Results are reported as averages across 10 seeds. ReCoRe 
 with segmentation (S) as a regularizer achieves state-of-the-art performance over competitors in the literature on 8 out of 12 experiments, where models are targeted for other (new) downstream tasks as well. Additionally, we report RAD \cite{data-aug-laskin20}, SAC+AE \cite{sac-ae-yarats2021}, and Pixel SAC \cite{p-or-s-sac-haarnoja2018} which do not consider additional downstream tasks. However, as explained in the literature review if an agent does not need to consider the downstream tasks then optimization of the additional constraints as in ReCoRe and CURL is not necessary, and thus performance is expected to improve (this is also noted by CURL \cite{curl-laskin20}). Thus ReCoRe, CURL and Dreamer form the main comparison, as these methods optimize the features learning and the controller separately. 
}
\label{tbl:dmc-results}
\end{table*}

We report the average SR and SPL on the held-out data in Table \ref{tbl:igibson1-results} after three training runs with random seeds. Our proposed ReCoRe outperforms state-of-the-art model-based RL method DreamerV2 and model-free methods RAD and CURL on 100k and 500k interactive steps. Since DreamerV2 is natively not trained with \textit{data augmentation} (DA), we include DreamerV2 + DA in our evaluation to show a fair comparison as all other methods contain DA inherently. We furthermore include transformer backed MWM and language guided foundation model DV2~
+~G~DINO. Nevertheless, the proposed ReCoRe still outperforms the competitors through the invariant features learning technique with \textit{contrastive loss} (CL) and \textit{regularization in the form of invariant auxiliary task} (D). Further, our results show that data augmentation can also improve model-based RL, which was previously shown only for model-free RL methods \cite{data-aug-laskin20}. 

\subsection{Sim-to-Real Transfer}
We use the Gibson dataset \cite{gibson-xia} for sim-to-real transfer experiments of the perception module, i.e. representation learning module of the world model; however please note that the robot controller is still a part of the simulator. Gibson scenes are created by 3D scanning real scenes, and a neural network is used to fill pathological geometric and occlusion errors. We have trained all models on the artist created textures of iGibson and tested on five scenes from Gibson. Table \ref{tbl:sim2real-gibson-results} shows the results. Our proposed ReCoRe significantly outperforms RAD, CURL and is slightly better compared to DV2+G DINO on 100k interactive steps, while on 500k steps ReCoRe significantly surpasses all the baselines. This shows that ReCoRe learns more stable features and is better suited for sim-to-real transfer.

\subsection{Generalizable Auxiliary Tasks}
The results for the DMControl suite \cite{dmc-tunyasuvunakool2020} experiments are shown in Table \ref{tbl:dmc-results}. We have used image denoising (I) and semantic segmentation (S) as the intervention invariant auxiliary tasks in these experiments. ReCoRe achieved competitive results, and the key findings are: i) even though depth reconstruction is an ideal task to enforce the invariant features learning on ReCoRe explicitly, the competitive results, even without depth reconstruction show the wider applicability of the proposed model; ii) ReCoRe outperform CURL on 8 out of 12 experiments, the closest state-of-the-art RL method with contrastive learning; iii) better invariant regularizer yields better results i.e. ReCoRe produces better results with segmentation than denoising as an auxiliary task.
Hence, we can conclude that ReCoRe is also competitive with end-to-end deep RL techniques even when training and evaluation environments come from similar distributions (and no OoD generalization is necessary).

Additionally we include other baseline models RAD \cite{data-aug-laskin20}, SAC+AE \cite{sac-ae-yarats2021}, and Pixel SAC \cite{p-or-s-sac-haarnoja2018}. However, as explained in the literature review, if an agent does not need to consider the downstream tasks then optimization of the additional constraints (as in ReCoRe and CURL) hinders the performance (also noted by CURL \cite{curl-laskin20}). So ReCoRe, CURL and Dreamer are the main competitors, as they optimize the feature learning and the controller separately. We have noticed that hyperparameter optimization for individual DMC tasks yields better results. However, we used the same set of parameters for all the tasks with ReCoRe.

\subsection{Ablation Study}
The contribution of contrastive learning and regularization in the form of intervention invariant auxiliary tasks are also shown in Table \ref{tbl:igibson1-results}. 
The standard formulation of contrastive learning does not use reconstruction loss \cite{simclr-chen20,byol-grill20,curl-laskin20,moco_he20}. Since model-based RL does not optimize the representation learning and controller jointly, contrastive loss collapses. Hence, to validate our proposal of intervention invariant depth reconstruction as a regularizer, we have done experiments without depth reconstruction (\mbox{ReCoRe - D}). We can see in Table \ref{tbl:igibson1-results} that in a reasonably complex pixel-based control task, ReCoRe is not able to learn meaningful control without the reconstruction task. Further, reconstructing RGB image (I) instead of depth (D), i.e. \mbox{ReCoRe - D + I}, slightly improves the results over \mbox{ReCoRe - D}, but is still approximately three times worse than the proposed ReCoRe. 

ReCoRe without contrastive loss (ReCoRe - CL) is unable to learn meaningful control. However, data augmentation (ReCoRe - CL + DA) slightly improves these results, but is still significantly worse than the competitors on OoD generalization. Hence, these results confirm that our proposal of doing an intervention on RGB observation space and adding intervention invariant reconstruction of depth as a regularizer is a crucial necessity for facilitating the proposed world model with invariant features.

\section{Conclusion}

We proposed a method to learn a \textit{World Model with invariant features}, ReCoRe. These invariant features are learned by minimizing contrastive loss between content invariance interventions of the observation. Hence, we proposed an auxiliary task as a regularizer, which is invariant to the proposed data augmentation techniques. ReCoRe significantly outperforms the state-of-the-art models on OoD generalization, sim-to-real transfer and sample efficiency measures. As such, ReCoRe is a new state of the art in model-based RL for sample efficiency in OoD generalization. Finally, we note that our framework can be applied to other tasks and the design of interventions and invariant auxiliary losses will become an interesting research problem.

{
    \small
    \bibliographystyle{ieeenat_fullname}
    \bibliography{main}
}

\clearpage
\setcounter{section}{0}
\setcounter{figure}{0}
\setcounter{table}{0}
\maketitlesupplementary

\section{Qualitative Results of ReCoRe}
\begin{figure}[htb]
\begin{minipage}{\textwidth}
\begin{center}
\fbox{\includegraphics[width=0.47\textwidth]{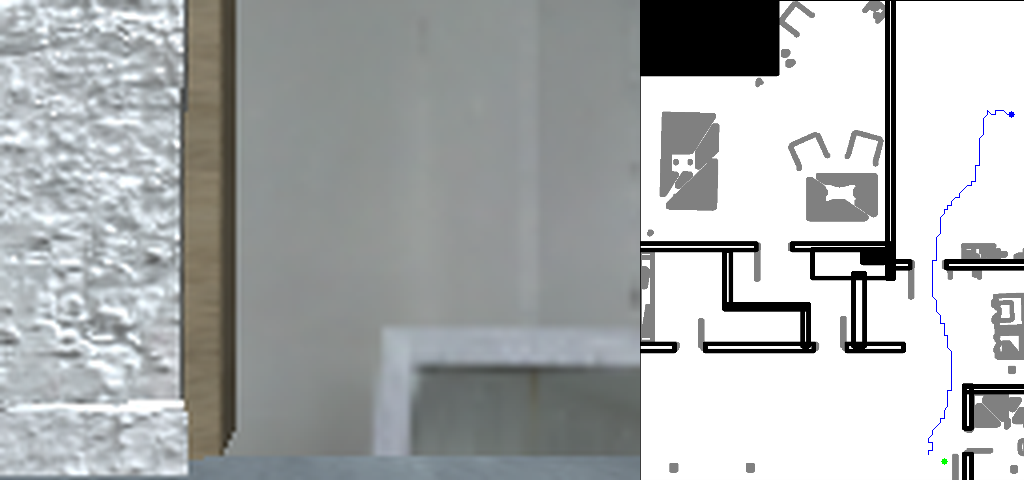}}
\fbox{\includegraphics[width=0.47\textwidth]{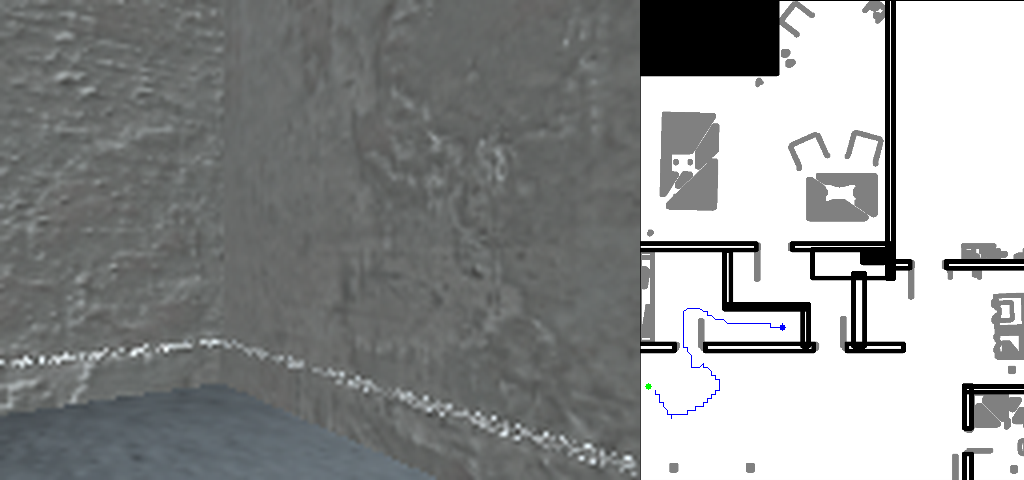}}
\fbox{\includegraphics[width=0.47\textwidth]{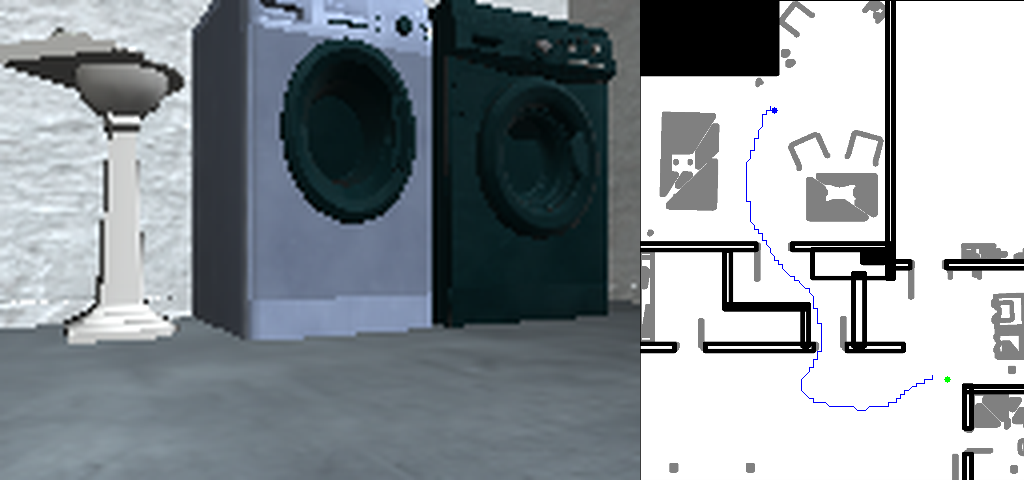}}
\fbox{\includegraphics[width=0.47\textwidth]{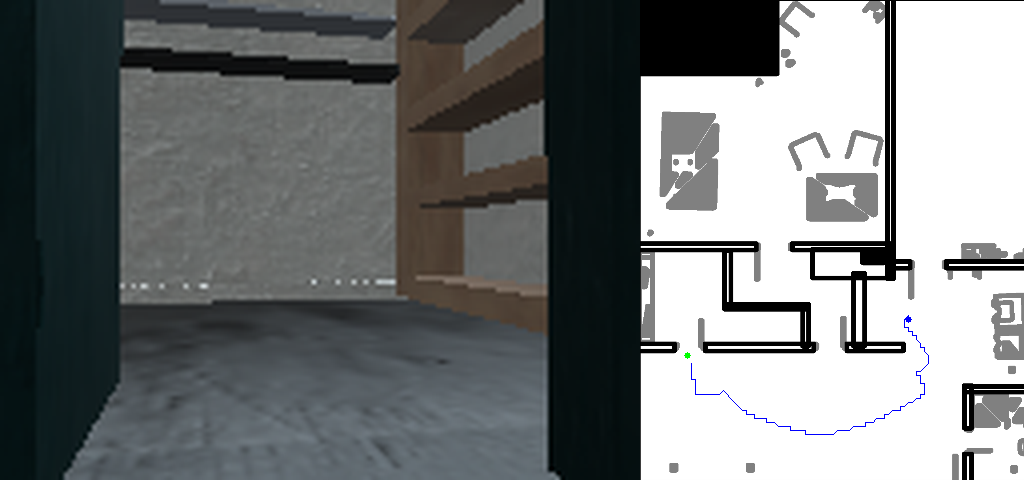}}
\fbox{\includegraphics[width=0.47\textwidth]{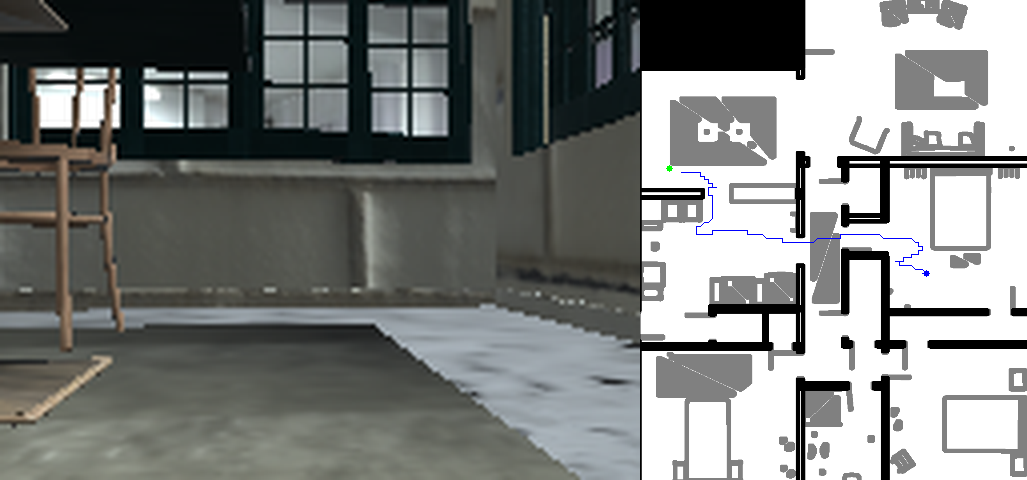}}
\fbox{\includegraphics[width=0.47\textwidth]{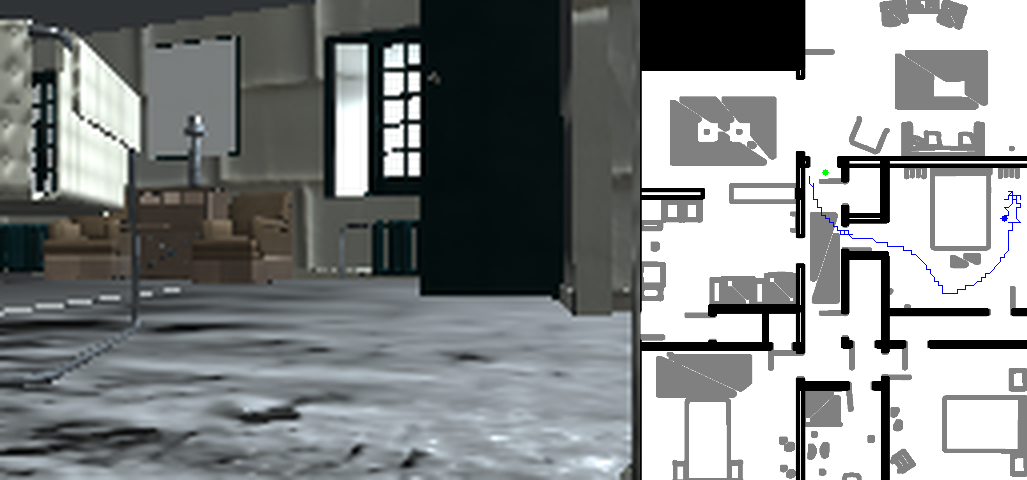}}
\end{center}
\caption{The out-of-distribution generalization tests of proposed ReCoRe on held-out scenes and visual textures from iGibson 1.0. Green circle is a random \textit{PointGoal}, blue circle is a random starting point and blue line represents the travel path of the Turtlebot robot.}
\label{fig:eg-more-travel-map.png}
\end{minipage}
\end{figure}
\vfill
\clearpage

\section{Hyper Parameters}
\begin{table}[htb]
\begin{minipage}{\textwidth}
\begin{center}
\begin{tabular}{lcc}
\hline
\noalign{\smallskip}
\textbf{Name} & \textbf{Symbol} & \textbf{Value} \\
\noalign{\smallskip}
\hline
\noalign{\smallskip}
World Model \\
\noalign{\smallskip}
\hline
\noalign{\smallskip}
Dataset size (FIFO) & --- & $3\cdot10^5$ \\
iGibson input image size & $o$ & 120$\times$160 \\
Batch size & $B$ & 50 \\
Sequence length & $L$ & 50 \\
Discrete latent dimensions & --- & 32 \\
Discrete latent classes & --- & 32 \\
RSSM number of units & --- & 1024 \\
KL loss scale & $\beta$ & 1.0 \\
World model learning rate & --- & $3\cdot10^{-4}$ \\
Key encoder exponential moving average & --- & 0.999 \\
\noalign{\smallskip}
\hline
\noalign{\smallskip}
Behavior \\
\noalign{\smallskip}
\hline
\noalign{\smallskip}
Imagination horizon & $H$ & 15 \\
Actor learning rate & --- & $1\cdot10^{-4}$ \\
Critic learning rate & --- & $1\cdot10^{-4}$ \\
Slow critic update interval & --- & $100$ \\
\noalign{\smallskip}
\hline
\noalign{\smallskip}
Common \\
\noalign{\smallskip}
\hline
\noalign{\smallskip}
Policy steps per gradient step & --- & 4 \\
Policy and reward MPL number of layers & --- & 4 \\
Policy and reward MPL number of units & --- & 400 \\
Gradient clipping & --- & 100 \\
Adam epsilon & $\epsilon$ & $10^{-5}$ \\
\noalign{\smallskip}
\hline
\noalign{\smallskip}
Encoder and Decoder \\
\noalign{\smallskip}
\hline
\noalign{\smallskip}
MLP encoder sizes of task obs & --- & 32, 32\\
Encoder kernels sizes & --- & 4, 4, 4, 4, 4\\
Decoder kernels sizes & --- & 5, 5, 4, 5, 4\\
Encoder and decoder feature maps & --- & 32, 64, 128, 256, 512\\
Encoder and decoder strides & --- & 2, 2, 2, 2, 2\\
Decoder padding & --- & none, 0-1, none, none, none\\
\noalign{\smallskip}
\hline
\noalign{\smallskip}
Data Augmentation \\
\noalign{\smallskip}
\hline
\noalign{\smallskip}
Padding range & --- & 10\\
Hue delta & --- & 0.1\\
Brightness delta & --- & 0.4\\
Contrast delta & --- & 0.4\\
Saturation delta & --- & 0.2\\
Gaussina blur sigma min, max & --- & 0.1, 2.0\\
Cutout min, max & --- & 30, 50\\
\noalign{\smallskip}
\hline
\noalign{\smallskip}
\end{tabular}
\end{center}
\caption{Hyper parameters of proposed ReCoRe.}
\label{tbl:recore-hparams}
\end{minipage}
\end{table}

\clearpage
\section{iGibson 1.0 Training and Evaluation Splits}
\begin{table}[htb]
\begin{minipage}{\textwidth}
\begin{center}
\begin{tabular}{lll}
\hline
\textbf{Phase} & & \textbf{Scene names} \\
\hline
Training & \hspace{1cm} & Beechwood\_0\_int, Beechwood\_1\_int, \\ 
         & \hspace{1cm} & Benevolence\_0\_int, Benevolence\_1\_int, Benevolence\_2\_int\\
         & \hspace{1cm} & Merom\_0\_int, Merom\_1\_int, \\
         & \hspace{1cm} & Pomaria\_0\_int, Pomaria\_1\_int, Pomaria\_2\_int,\\
         & \hspace{1cm} & Wainscott\_0\_int, Wainscott\_1\_int \\
Testing & \hspace{1cm} & Ihlen\_0\_int, Ihlen\_1\_int, Rs\_int\\
\hline
\end{tabular}
\end{center}
\caption{Train-test scenes splits for iGibsion 1.0 dataset. }
\label{tbl:igibson1-0-splits}
\end{minipage}
\end{table}

\begin{table}[htb]
\begin{minipage}{\textwidth}
\begin{center}
\begin{tabular}{lll}
\hline
\textbf{Material category} & \hspace{1.5cm} & \textbf{Held-out texture ids for test}\\
\hline
asphalt               && 06, 15             \\
bricks                && 08, 19             \\
concrete              && 06, 15, 17         \\
fabric                && 01, 02, 28         \\
fabric\_carpet        && 02, 05, 13         \\
ground                && 13, 19             \\
leather               && 03, 12             \\
marble                && 02, 03             \\
metal                 && 10, 19             \\
metal\_diamond\_plate && 04                 \\
moss                  && 01, 03             \\
paint                 && 05                 \\
paving\_stones        && 24, 38             \\
planks                && 07, 09, 16         \\
plaster               && 03                 \\
plastic               && 04, 05             \\
porcelain             && 02, 04             \\
rocks                 && 04                 \\
terrazzo              && 06, 08             \\
tiles                 && 43, 49             \\
wood                  && 02, 05, 16, 22, 32 \\
wood\_floor           && 06, 10, 17, 28     \\ 
\hline
\end{tabular}
\end{center}
\caption{iGibson 1.0 environment held out texture ids for test.}
\label{tbl:held-out-textures}
\end{minipage}

\end{table}
\clearpage


\begin{table}[htb]
\begin{minipage}{\textwidth}
\begin{center}
\begin{tabular}{ccccc}
\hline
\noalign{\smallskip}
\multirow{2}{*}{concrete} & \multirow{1}{*}{Train} & \includegraphics[width=0.16\textwidth,height=0.12\textwidth]{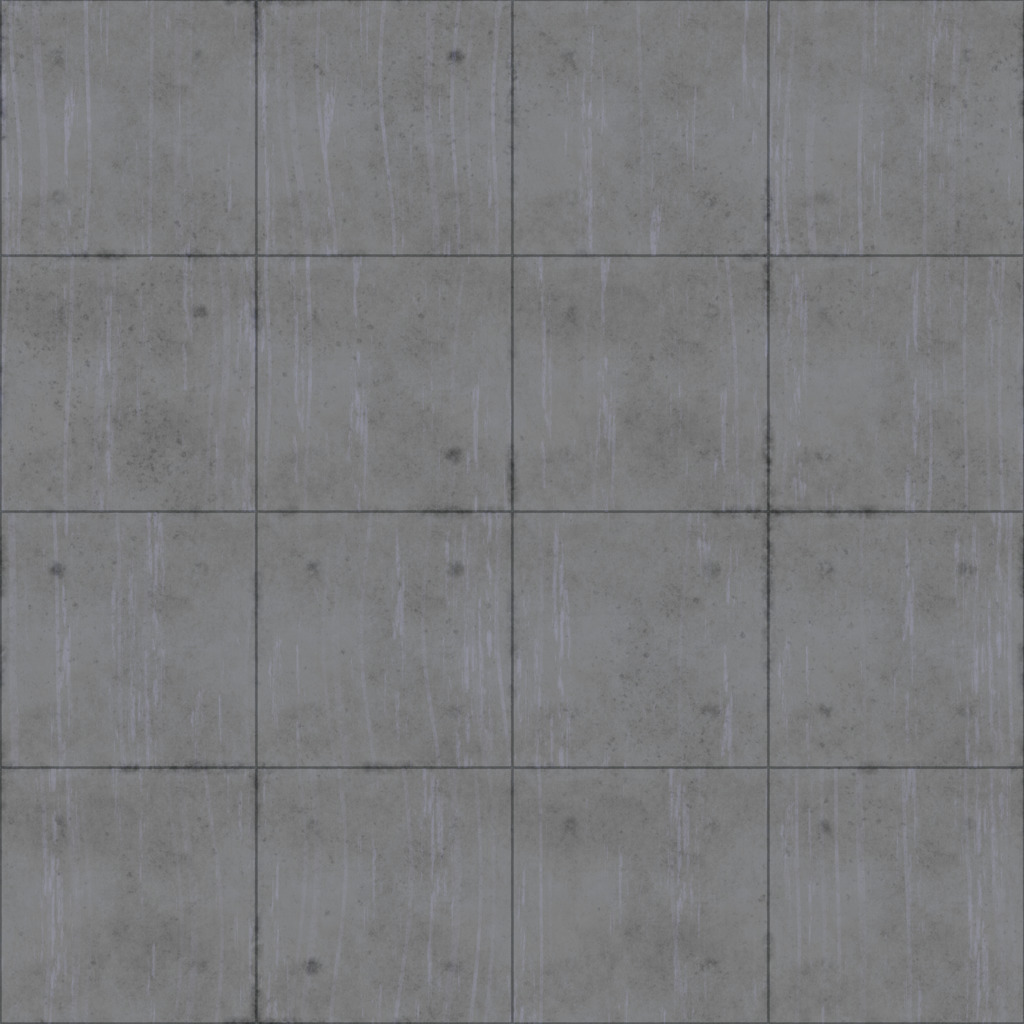} & 
                                    \includegraphics[width=0.16\textwidth,height=0.12\textwidth]{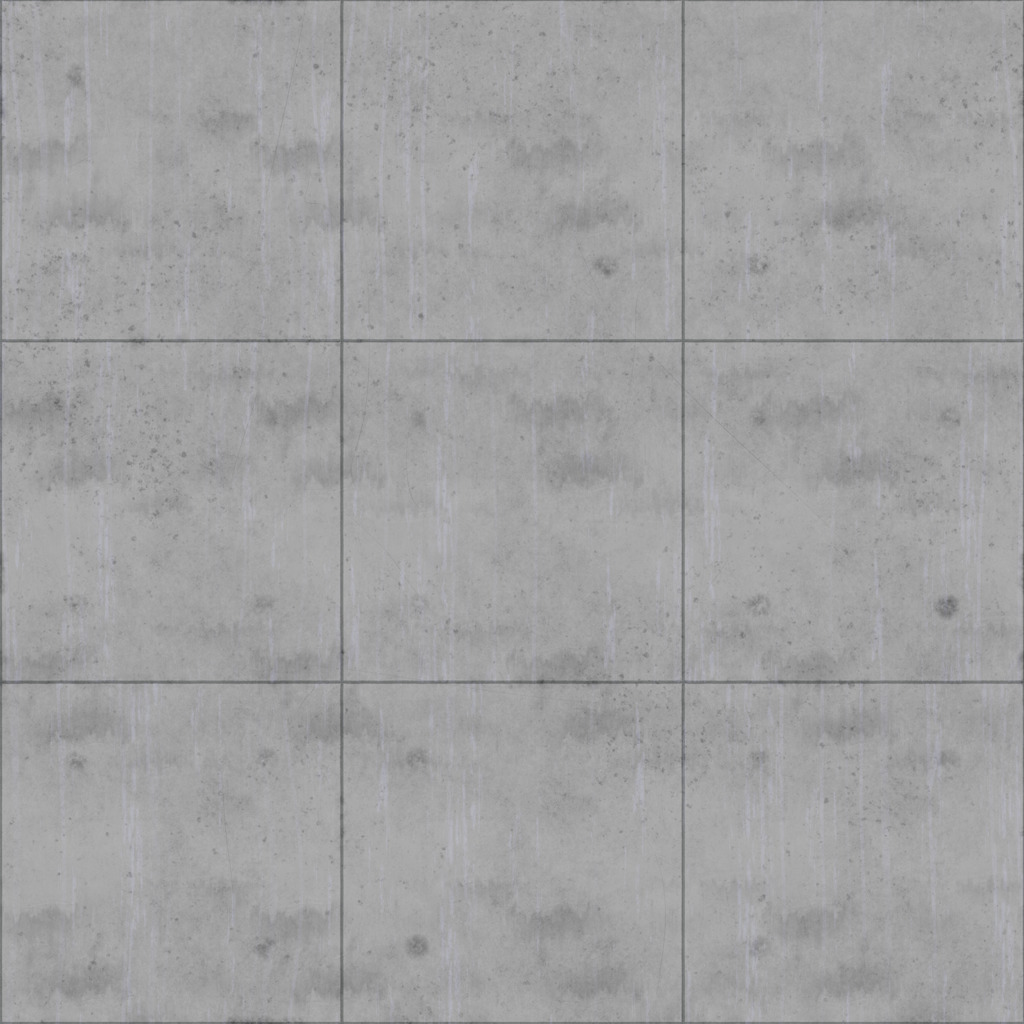} & 
                                    \includegraphics[width=0.16\textwidth,height=0.12\textwidth]{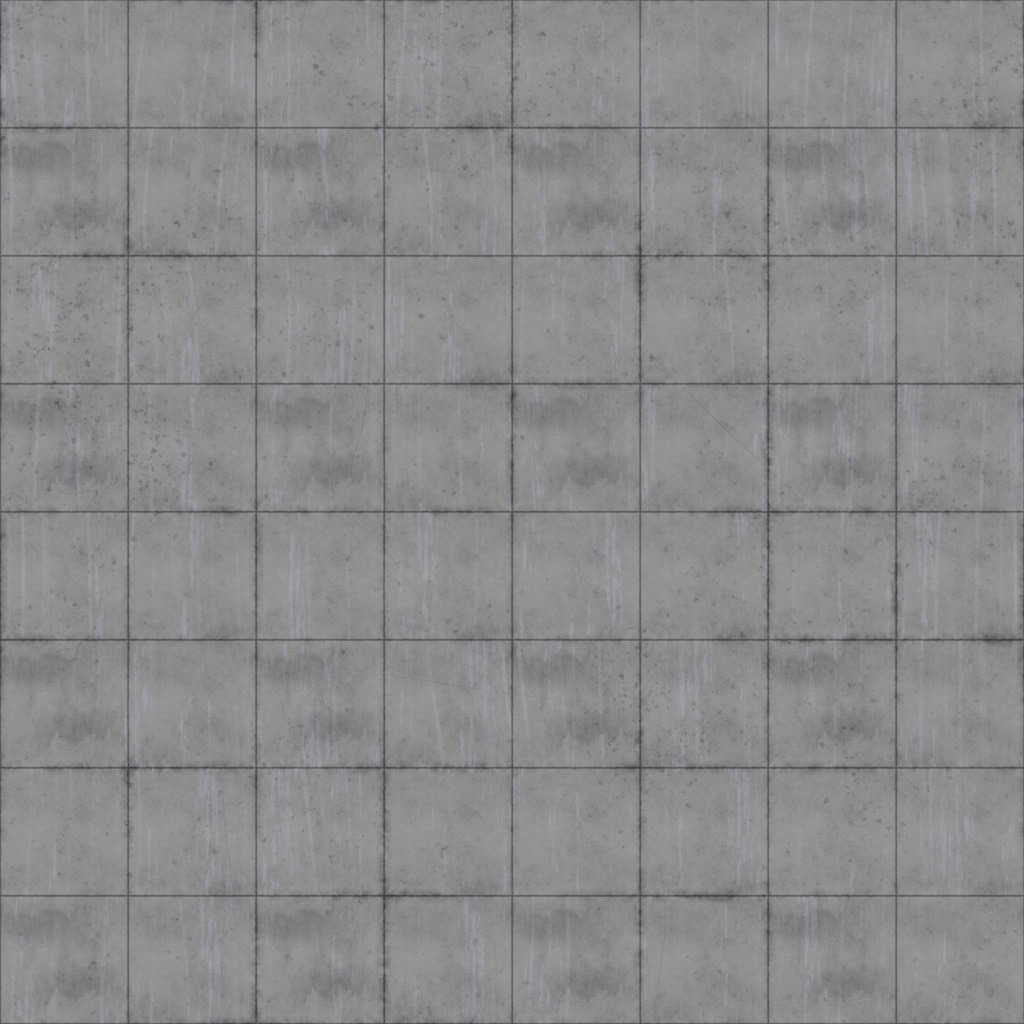} \\
                          
                          & \multirow{1}{*}{Test}  & \includegraphics[width=0.16\textwidth,height=0.12\textwidth]{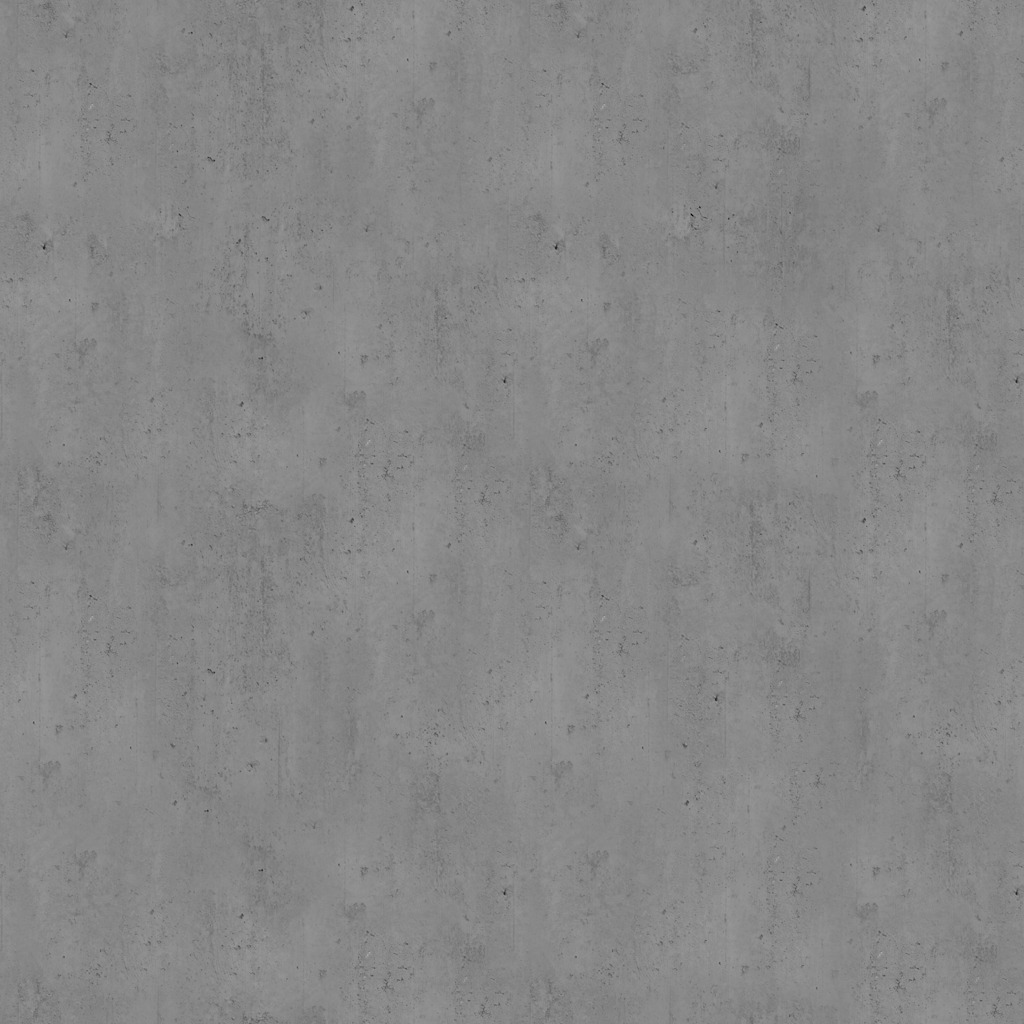} & 
                                    \includegraphics[width=0.16\textwidth,height=0.12\textwidth]{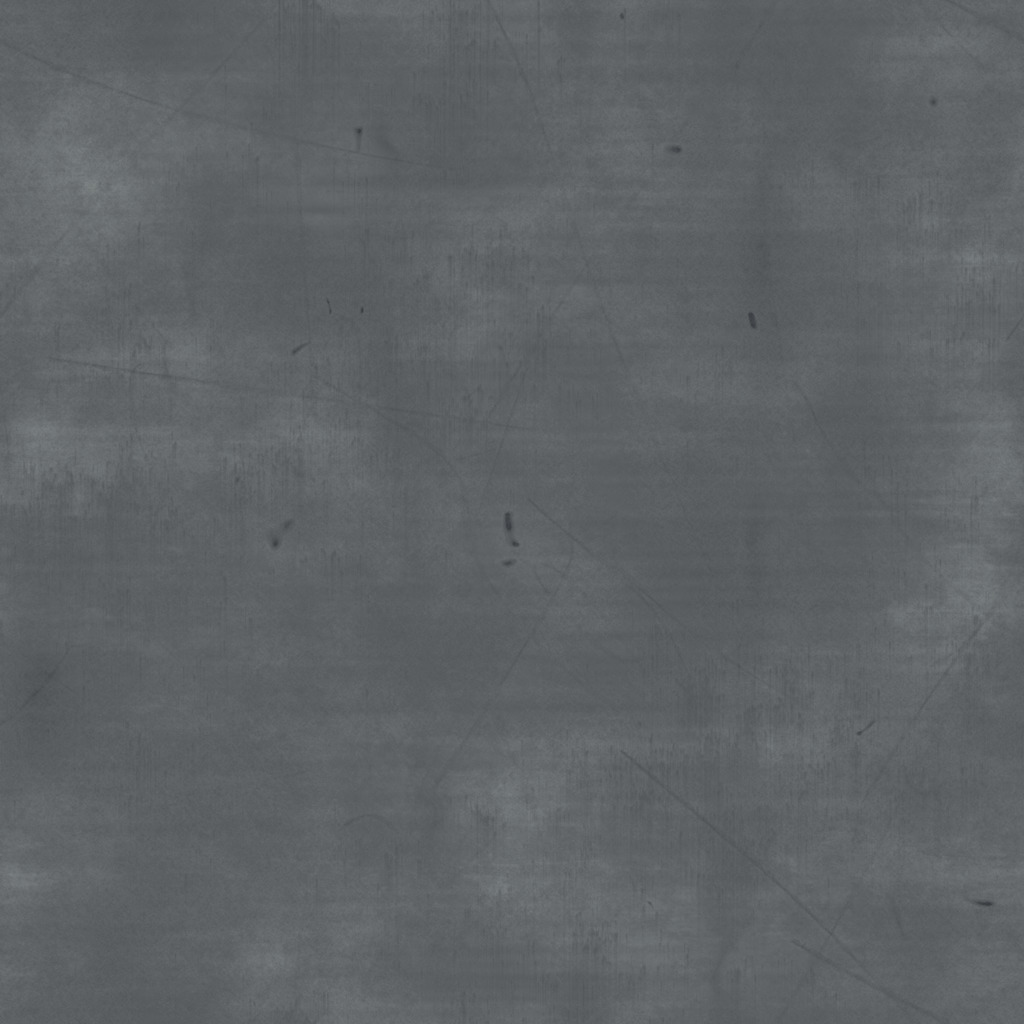} & 
                                    \includegraphics[width=0.16\textwidth,height=0.12\textwidth]{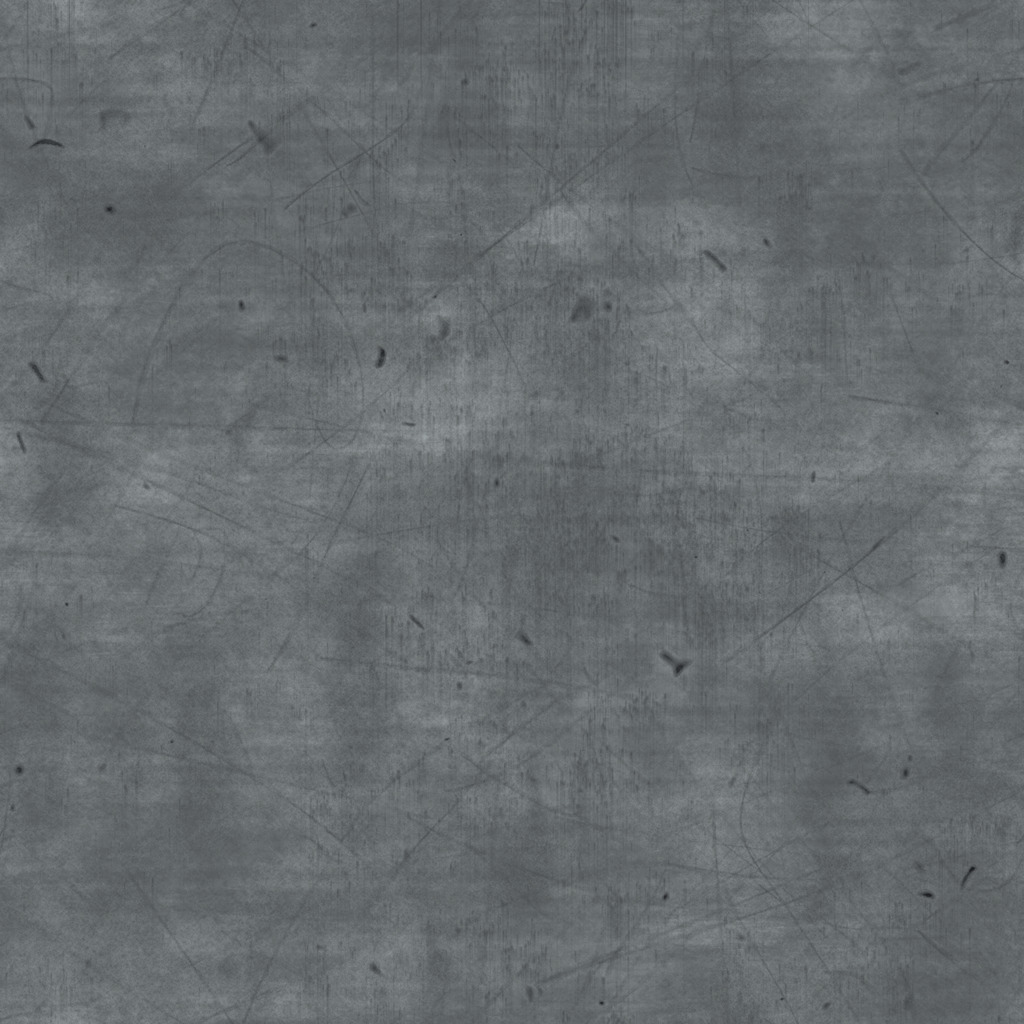} \\
\noalign{\smallskip}
\hline
\noalign{\smallskip}
\multirow{2}{*}{fabric}   & Train & \includegraphics[width=0.16\textwidth,height=0.12\textwidth]{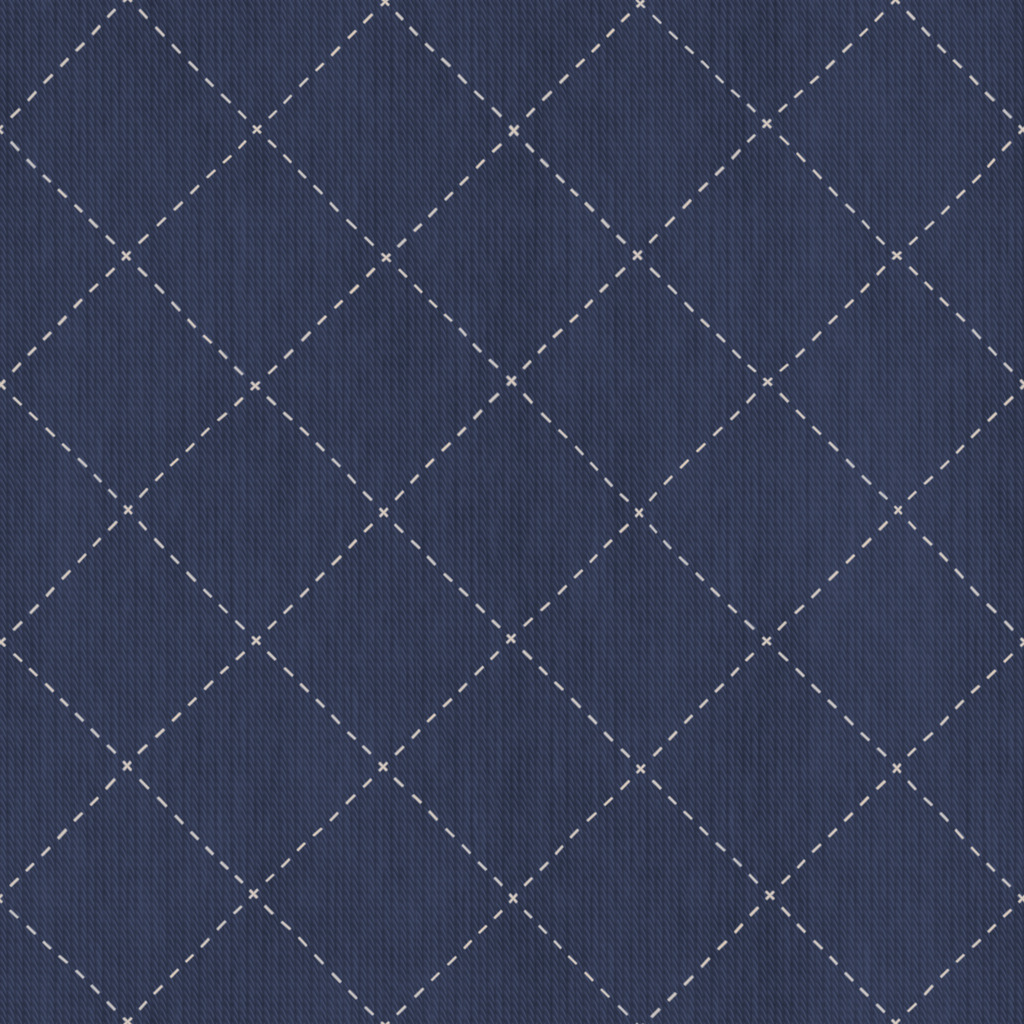} & 
                                    \includegraphics[width=0.16\textwidth,height=0.12\textwidth]{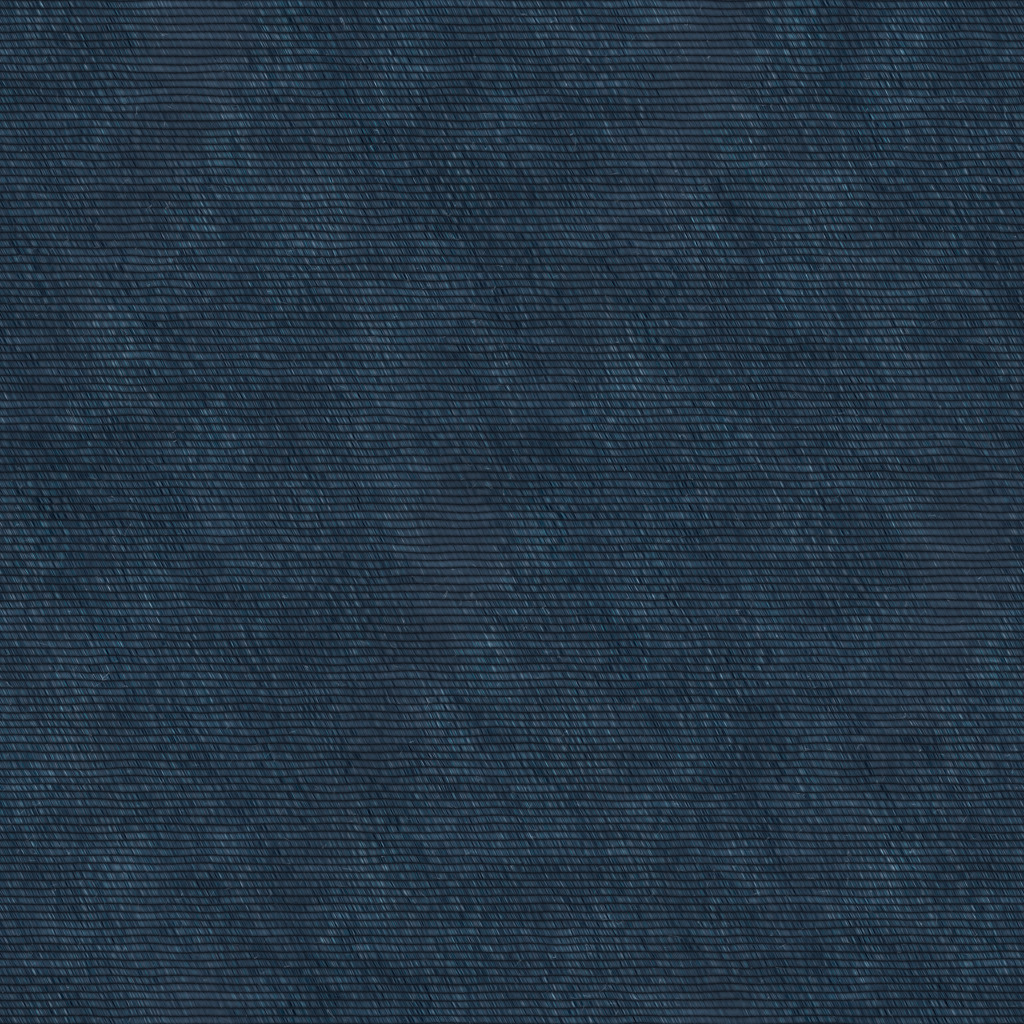} & 
                                    \includegraphics[width=0.16\textwidth,height=0.12\textwidth]{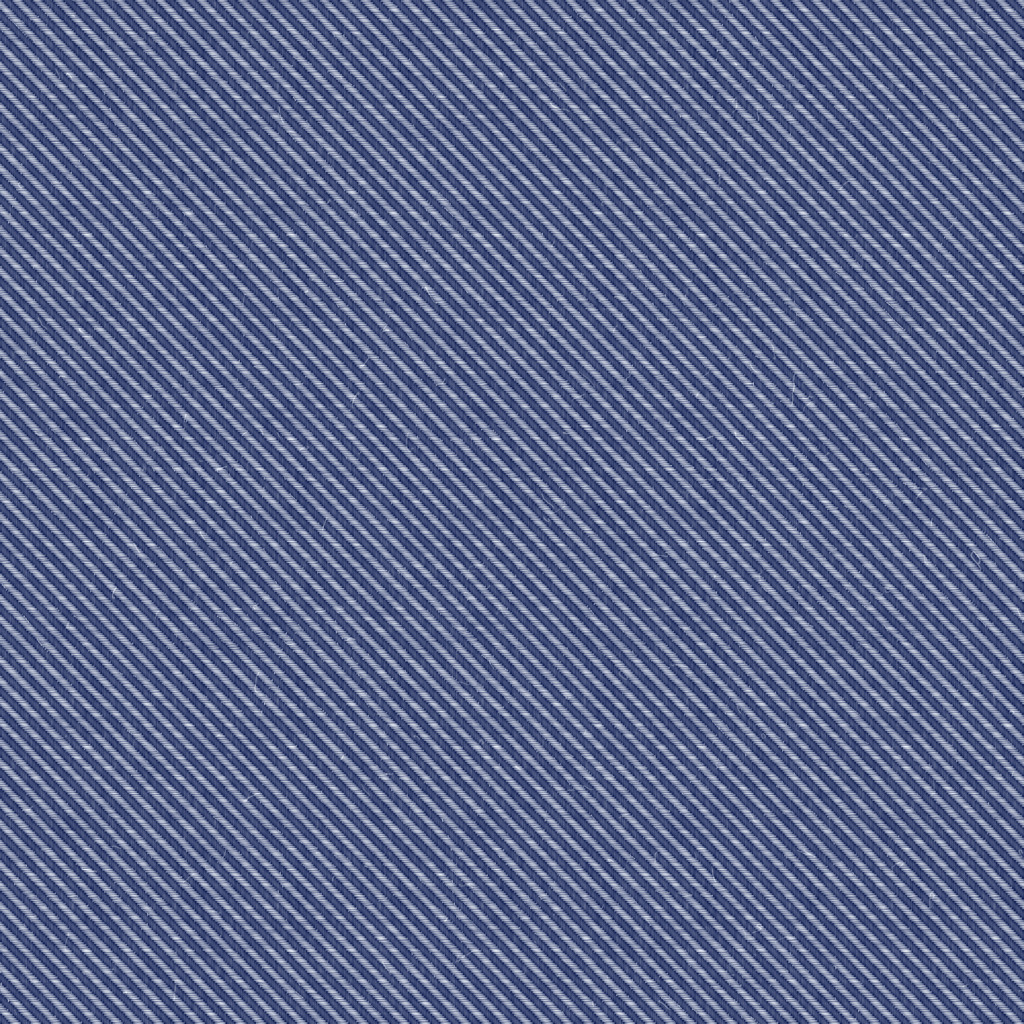} \\
                          
                          & Test  & \includegraphics[width=0.16\textwidth,height=0.12\textwidth]{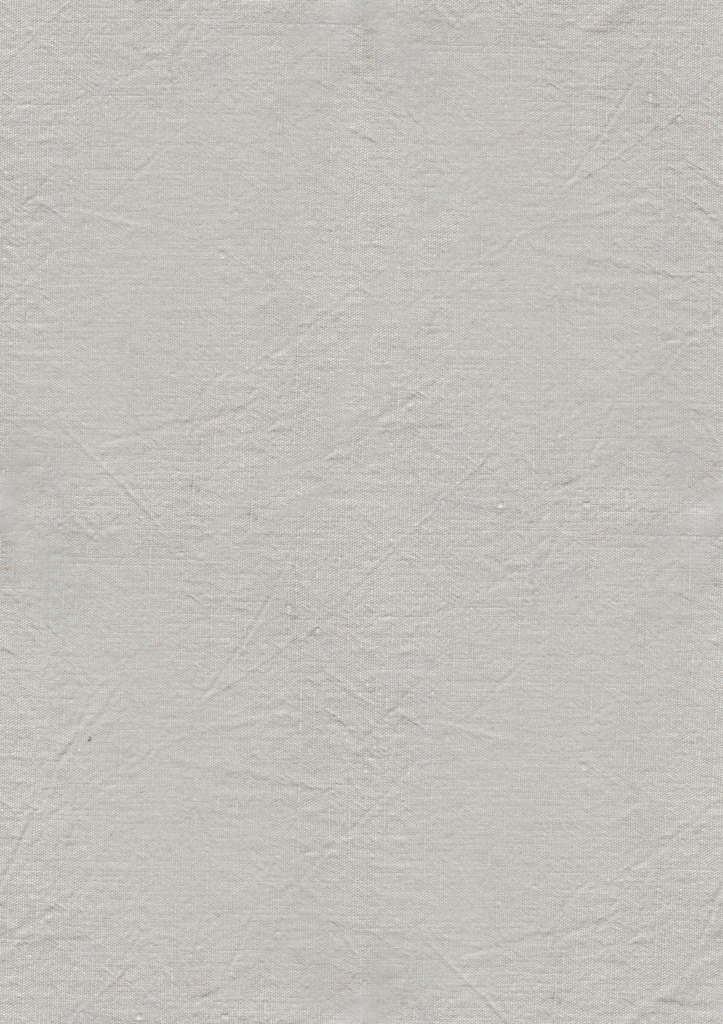} & 
                                    \includegraphics[width=0.16\textwidth,height=0.12\textwidth]{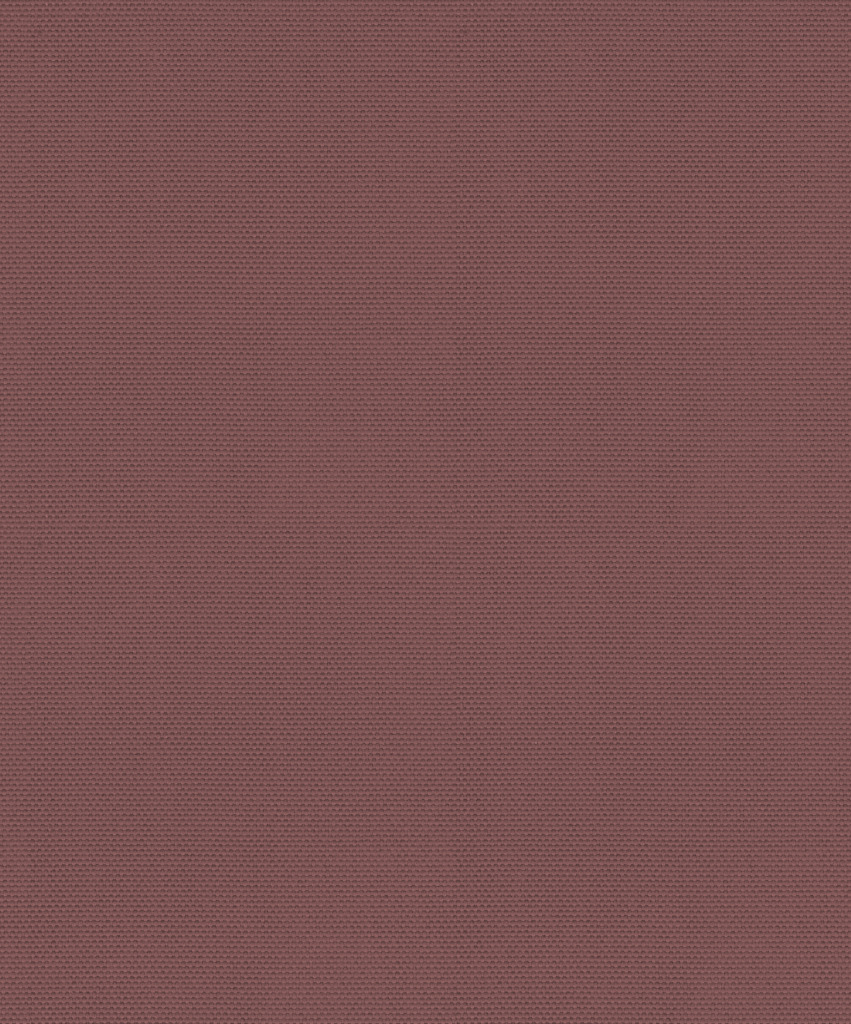} & 
                                    \includegraphics[width=0.16\textwidth,height=0.12\textwidth]{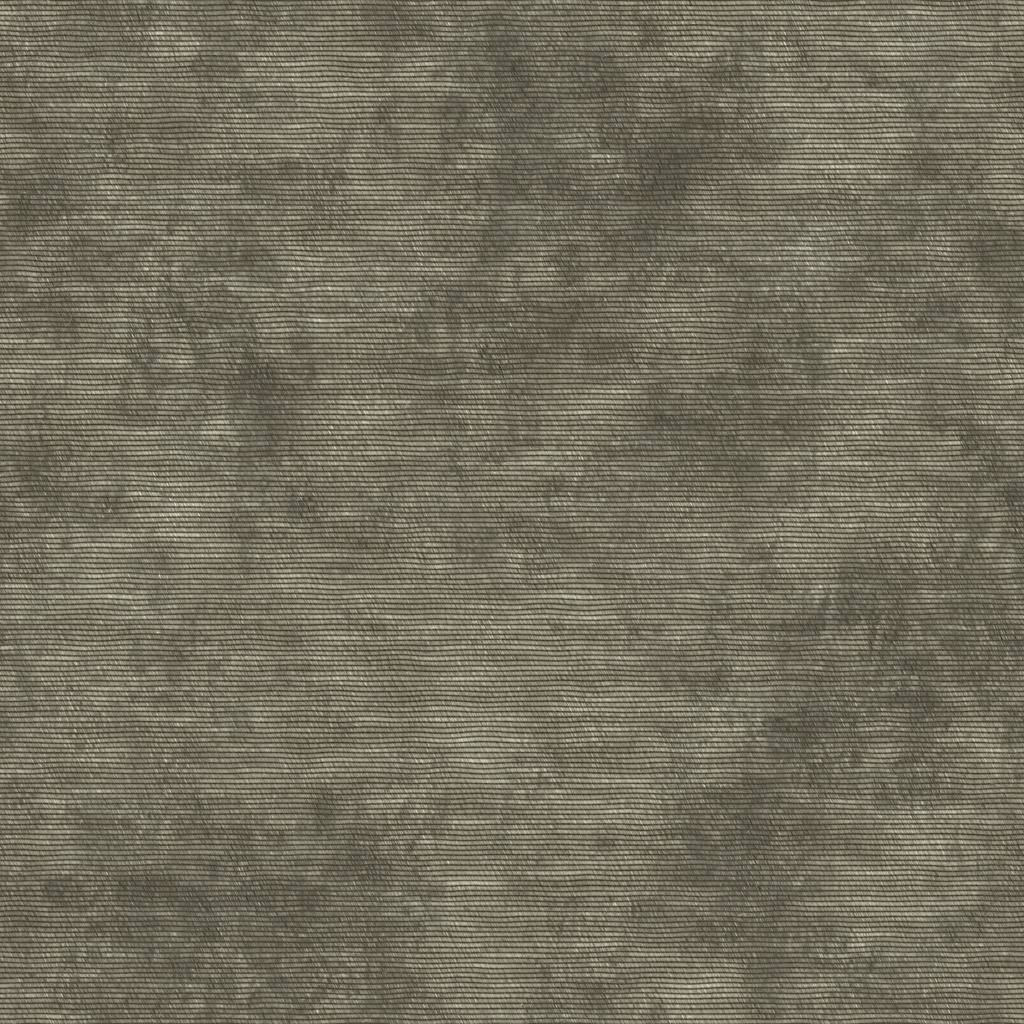} \\
\noalign{\smallskip}
\hline
\noalign{\smallskip}
\multirow{2}{*}{planks}   & Train & \includegraphics[width=0.16\textwidth,height=0.12\textwidth]{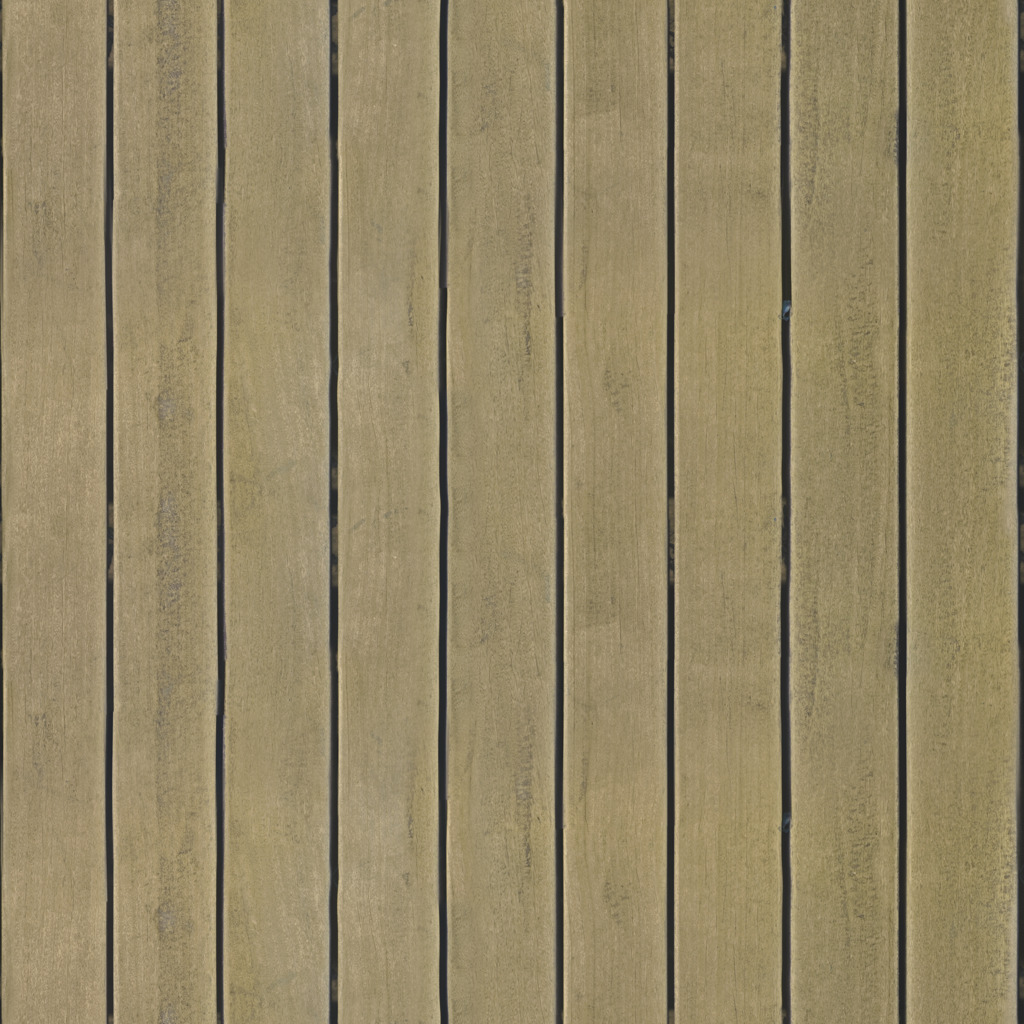} & 
                                    \includegraphics[width=0.16\textwidth,height=0.12\textwidth]{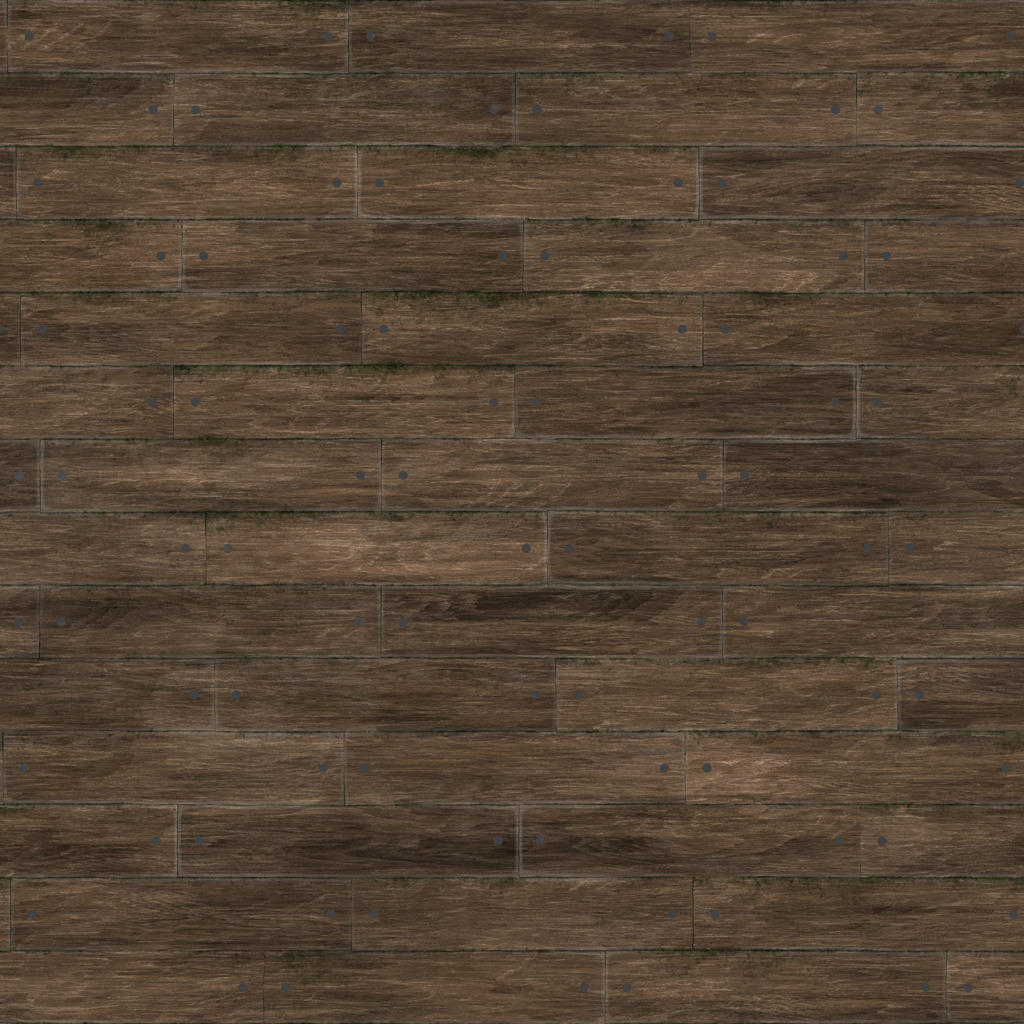} & 
                                    \includegraphics[width=0.16\textwidth,height=0.12\textwidth]{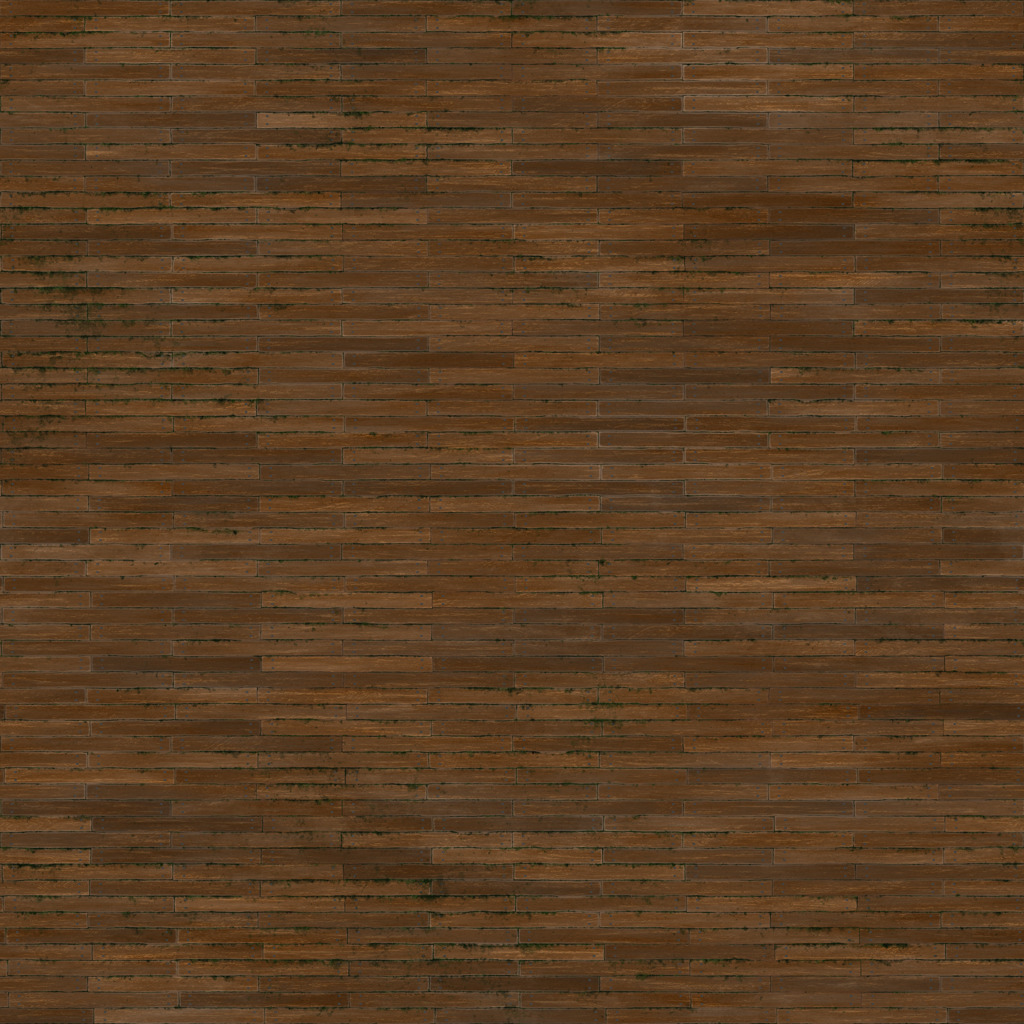} \\
                          
                          & Test  & \includegraphics[width=0.16\textwidth,height=0.12\textwidth]{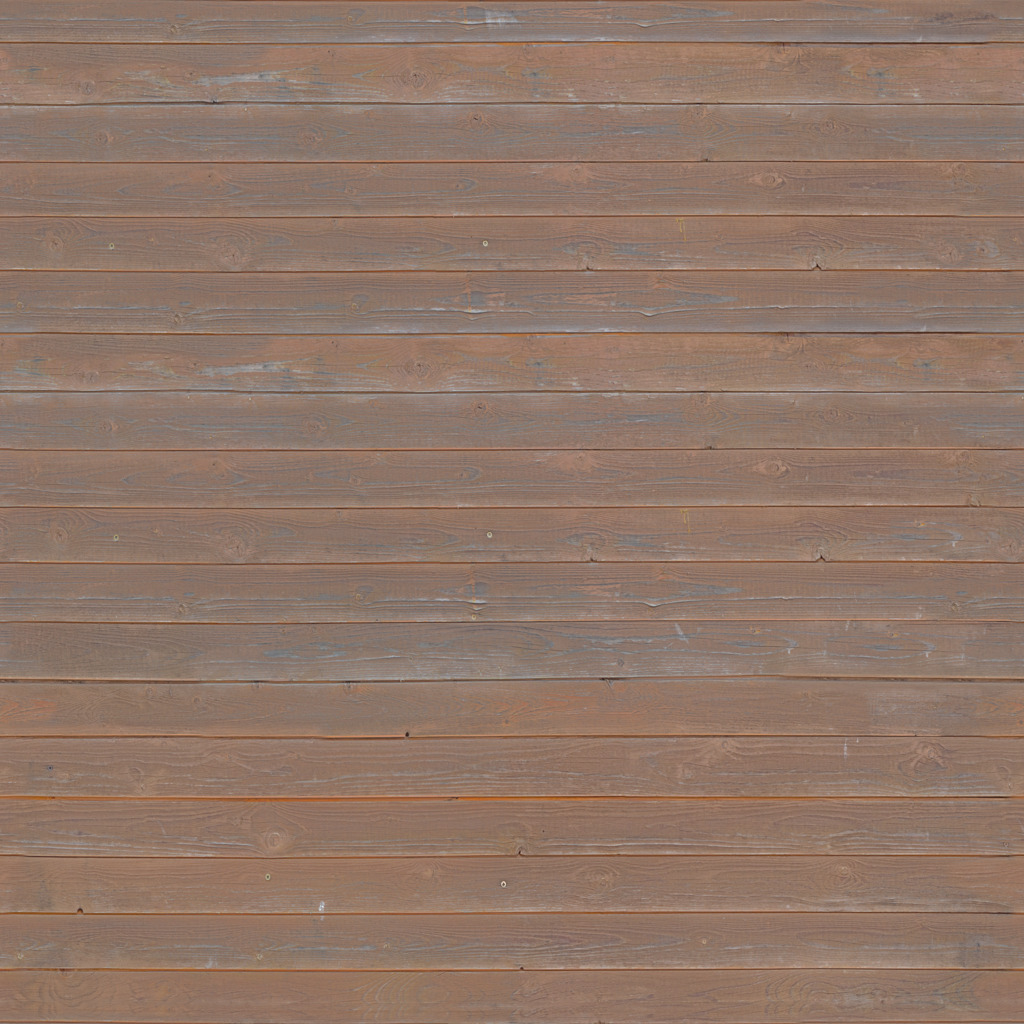} & 
                                    \includegraphics[width=0.16\textwidth,height=0.12\textwidth]{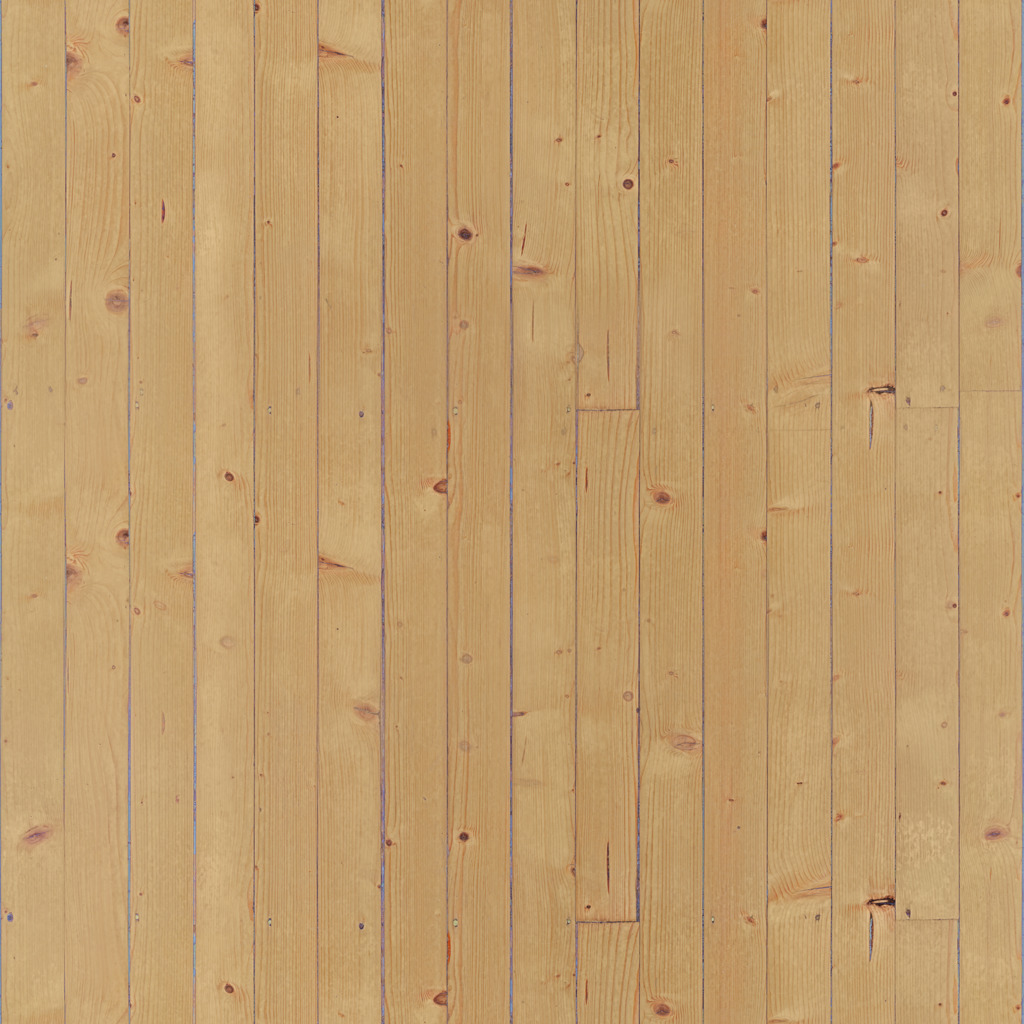} & 
                                    \includegraphics[width=0.16\textwidth,height=0.12\textwidth]{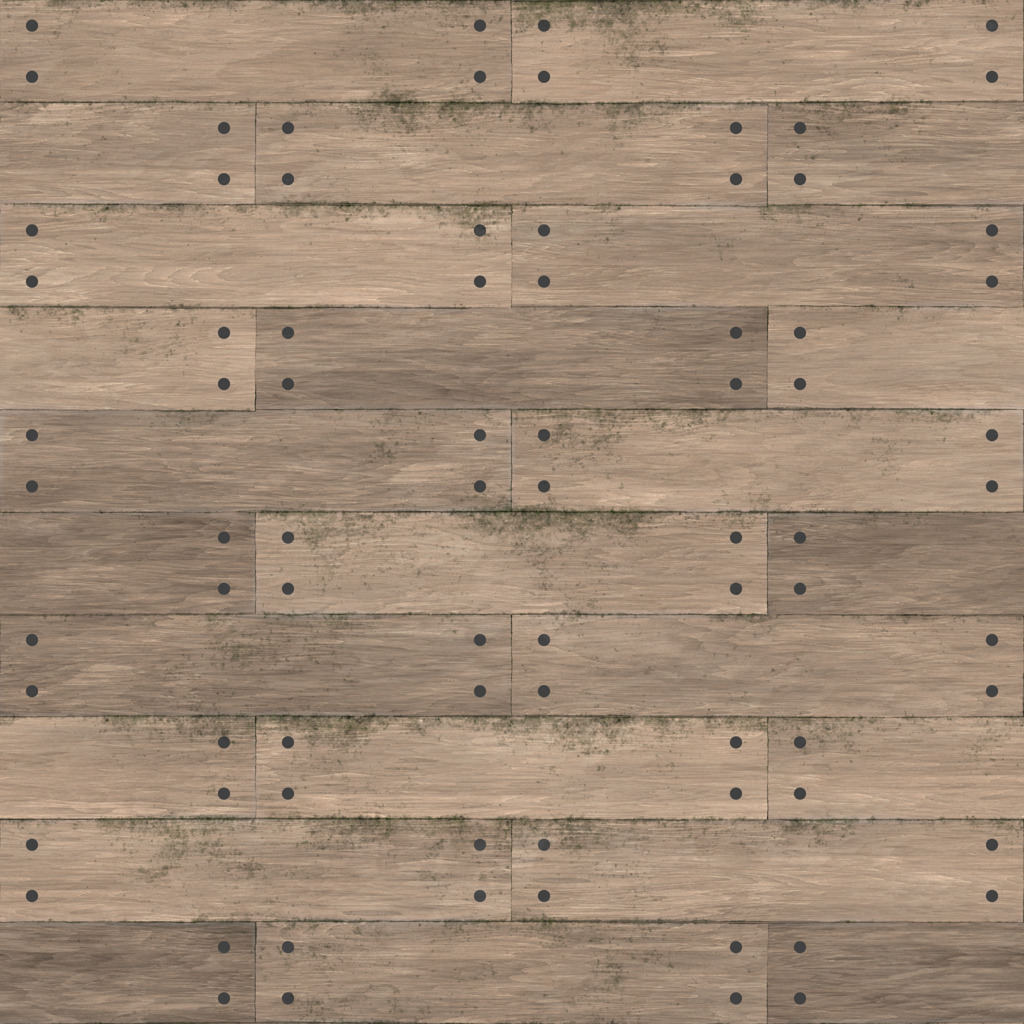} \\
\noalign{\smallskip}
\hline
\noalign{\smallskip}
\multirow{2}{*}{wood}     & Train & \includegraphics[width=0.16\textwidth,height=0.12\textwidth]{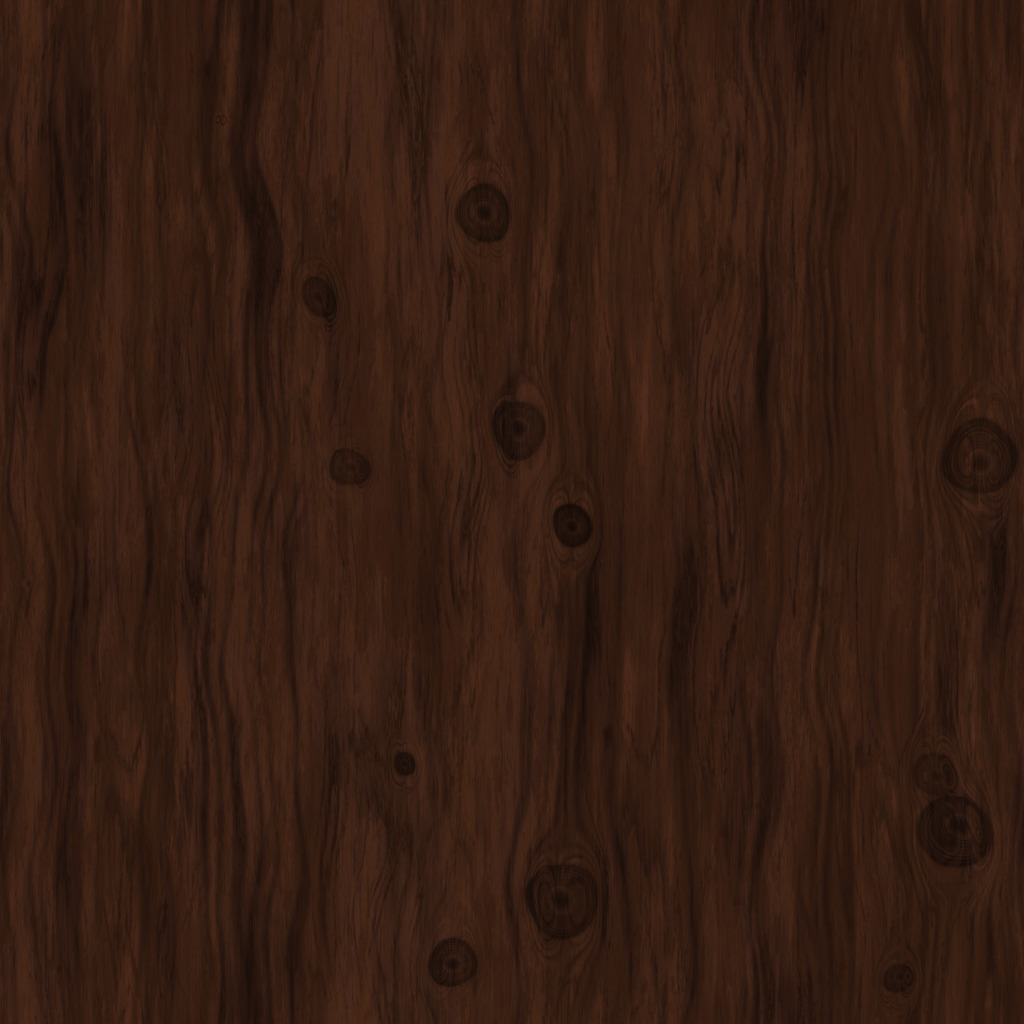} & 
                                    \includegraphics[width=0.16\textwidth,height=0.12\textwidth]{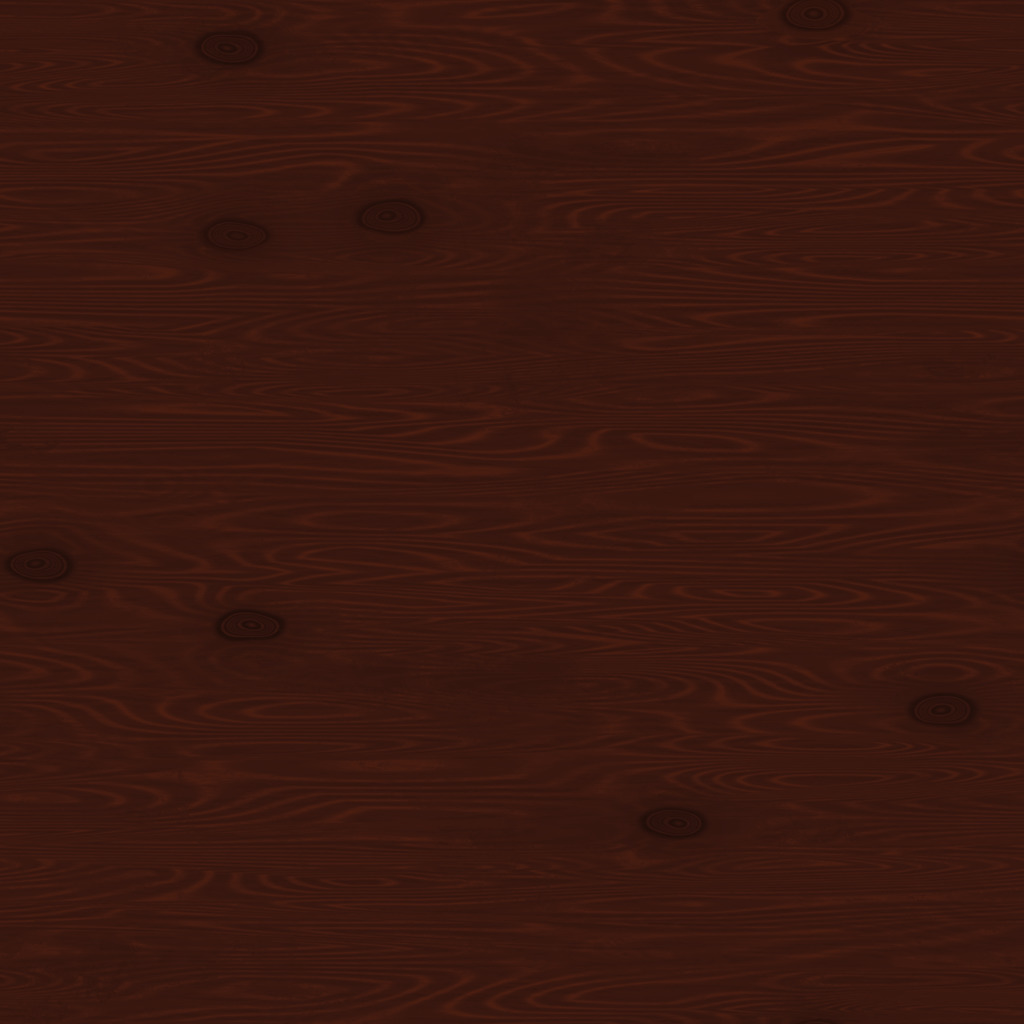} & 
                                    \includegraphics[width=0.16\textwidth,height=0.12\textwidth]{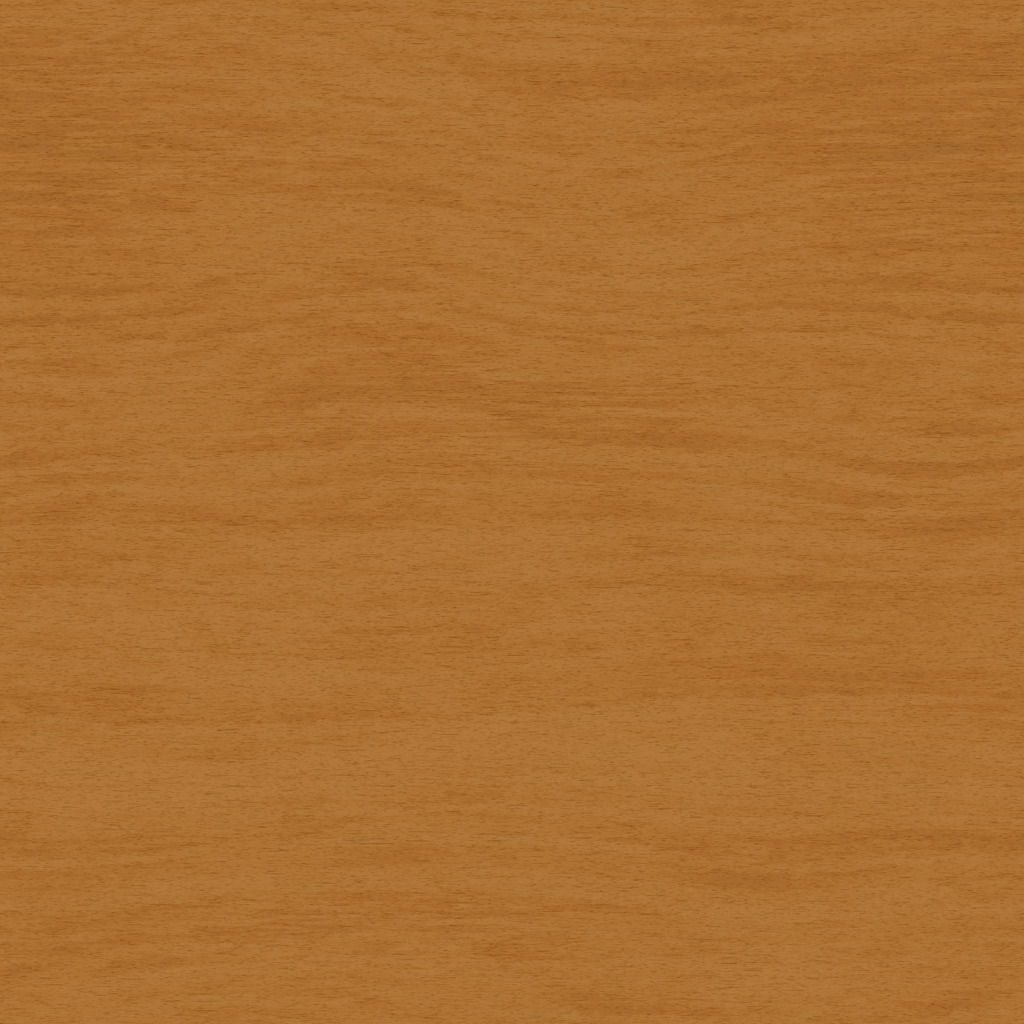} \\
                          
                          & Test  & \includegraphics[width=0.16\textwidth,height=0.12\textwidth]{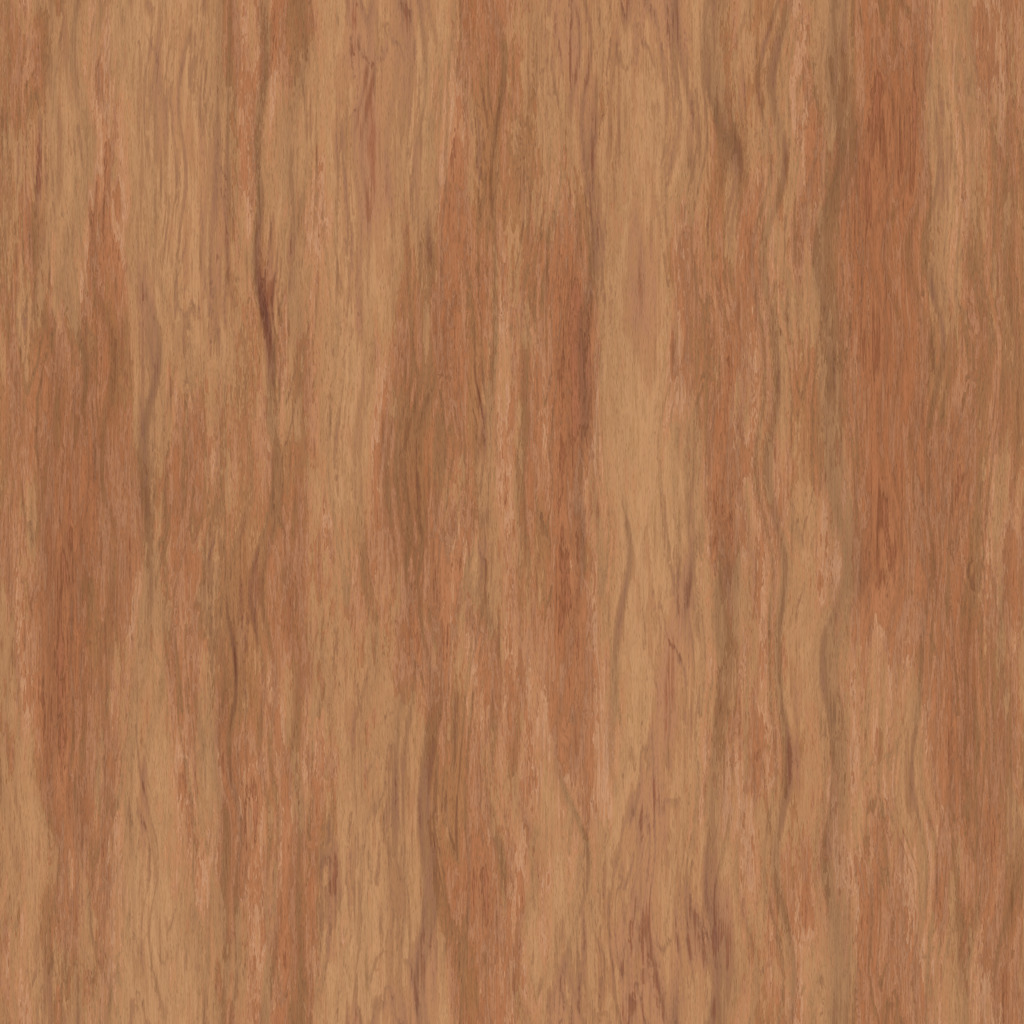} & 
                                    \includegraphics[width=0.16\textwidth,height=0.12\textwidth]{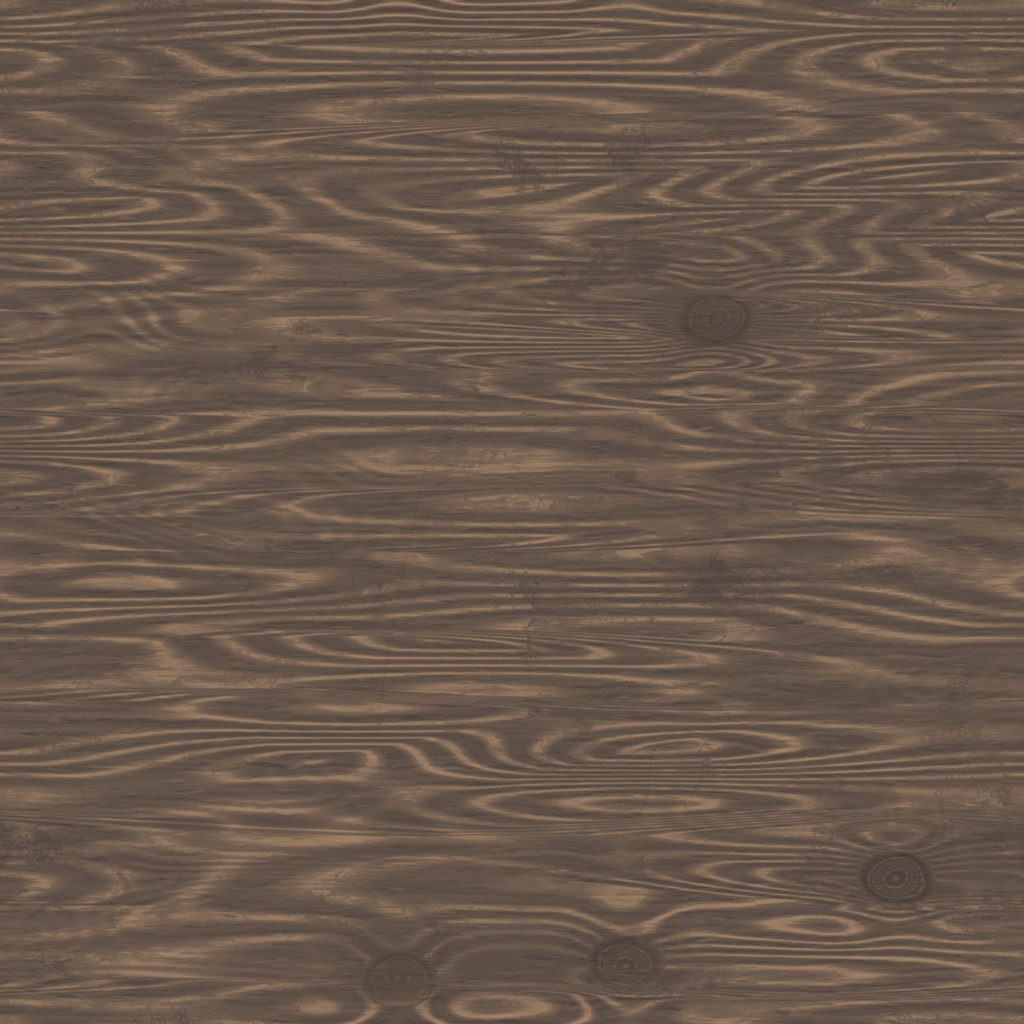} & 
                                    \includegraphics[width=0.16\textwidth,height=0.12\textwidth]{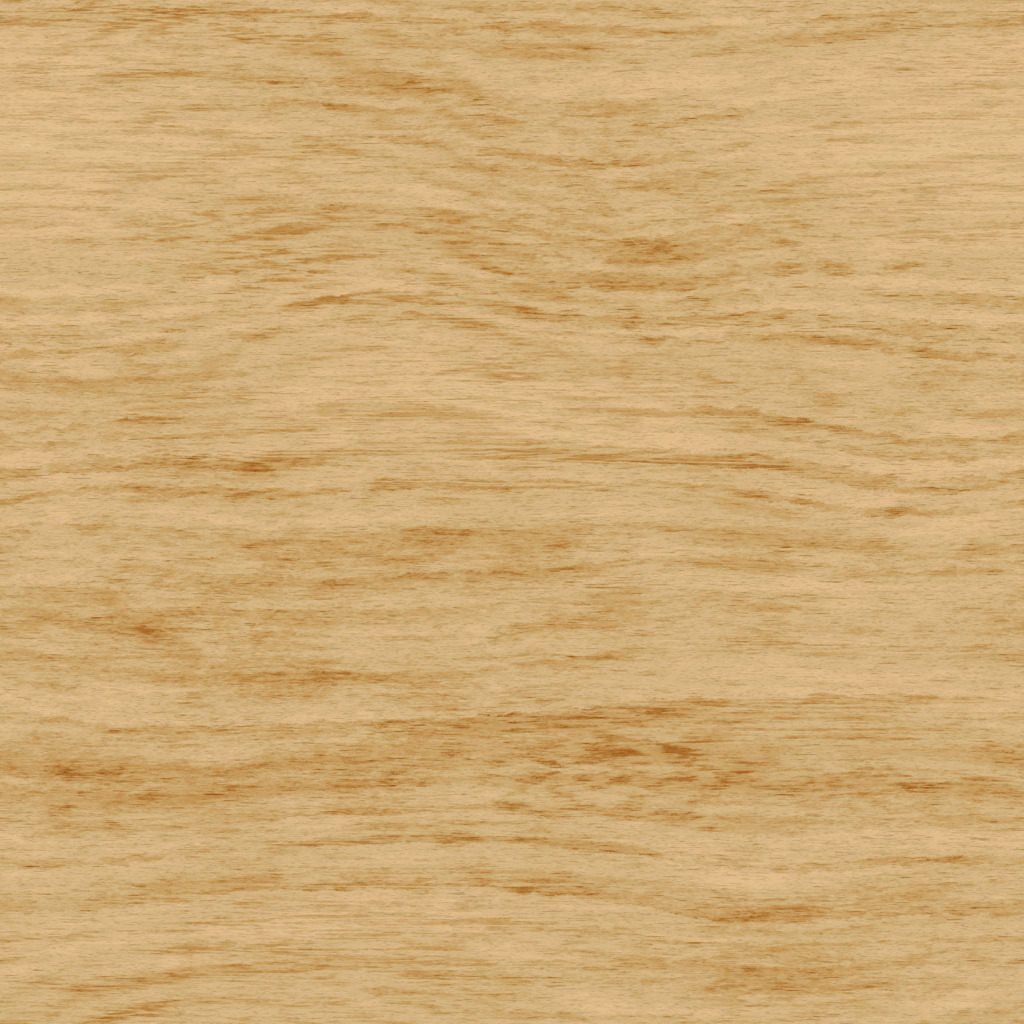} \\
\noalign{\smallskip}
\hline
\noalign{\smallskip}
\end{tabular}
\end{center}
\caption{Examples of textures from the training and testing splits of iGibson 1.0 benchmark. Note the significant variation between textures used in training and those employed in testing, which facilitates the evaluation of out-of-distribution generalization.}  

\label{tbl:held-out-textures-examples}
\end{minipage}
\end{table}

\end{document}

%% file: preamble.tex
\usepackage[dvipsnames]{xcolor}
